\documentclass[10pt,twocolumn,letterpaper]{article}

\usepackage{wacv}
\usepackage{times}
\usepackage{epsfig}
\usepackage{graphicx}
\usepackage{amsmath}
\usepackage{amssymb}
\usepackage{comment}
\usepackage{multirow}
\usepackage{subcaption}
\usepackage{cuted}
\usepackage{capt-of}
\usepackage[table]{xcolor}
\usepackage{array,float,xfrac}
\usepackage[accsupp]{axessibility}  
 
\usepackage{import}
\newcommand{\supplementarysection}{
	\setcounter{figure}{0}
	\setcounter{table}{0}
	\setcounter{section}{0}
	\renewcommand{\thesection}{S\arabic{section}}  
	\renewcommand{\thetable}{S\arabic{table}}  
	\renewcommand{\thefigure}{S\arabic{figure}}
}

\def\wacvPaperID{255} 

\wacvapplicationstrack 

\wacvfinalcopy 

\ifwacvfinal
\def\assignedStartPage{1} 
\fi

\ifwacvfinal
\usepackage[breaklinks=true,bookmarks=false]{hyperref}
\else
\usepackage[pagebackref=true,breaklinks=true,colorlinks,bookmarks=false]{hyperref}
\fi

\ifwacvfinal
\else
\fi

\begin{document}

\title{\vspace{-1cm}OpenEarthMap: \\A Benchmark Dataset for Global High-Resolution Land Cover Mapping\vspace{-.4cm}}

\author{Junshi Xia$^{1,}$\thanks{Equal contribution. $^{\dag}$Corresponding author.} , Naoto Yokoya$^{2,1,*,\dag}$, Bruno Adriano$^{1,*}$, and Clifford Broni-Bediako$^{1}$\\
$^{1}$RIKEN AIP, Japan {\tt\small \{junshi.xia,bruno.adriano,clifford.broni-bediako\}@riken.jp}\\
$^{2}$The University of Tokyo, Japan {\tt\small yokoya@k.u-tokyo.ac.jp}\\
}
\maketitle

\begin{strip}\centering
\vspace{-1.5cm} 
\includegraphics[width= \linewidth]{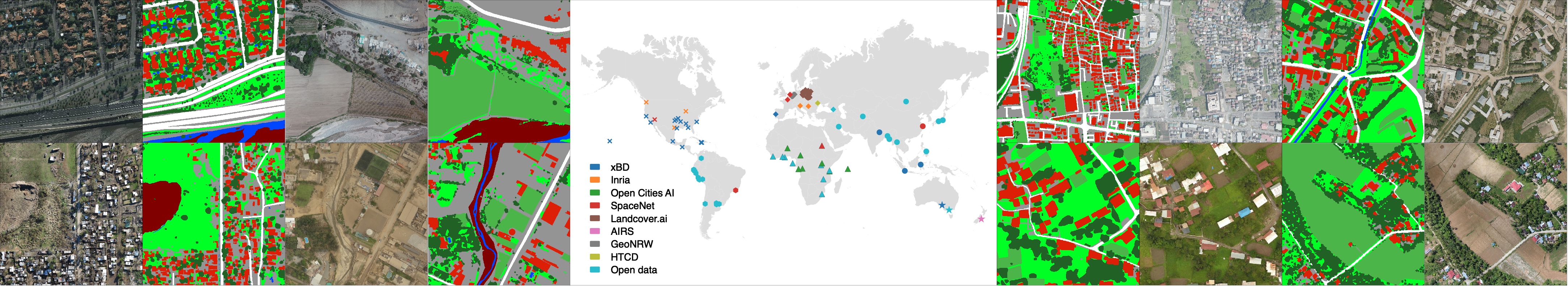}
\vspace{-0.55cm}
\captionof{figure}{A world map showing the locations of 97 regions included in OpenEarthMap and eight annotated examples.
\label{fig:feature-graphic}}
\vspace{-0.25cm}
\end{strip}

\begin{abstract}
\vspace{-0.3cm}
   We introduce OpenEarthMap, a benchmark dataset, for global high-resolution land cover mapping. OpenEarthMap consists of 2.2 million segments of 5000 aerial and satellite images covering 97 regions from 44 countries across 6 continents, with manually annotated 8-class land cover labels at a 0.25--0.5m ground sampling distance. Semantic segmentation models trained on the OpenEarthMap generalize worldwide and can be used as off-the-shelf models in a variety of applications. We evaluate the performance of state-of-the-art methods for unsupervised domain adaptation and present challenging problem settings suitable for further technical development. We also investigate lightweight models using automated neural architecture search for limited computational resources and fast mapping. The dataset is available at \url{https://open-earth-map.org}.
\end{abstract}

\begin{table*}[t!]
\caption{Summary of remote sensing benchmark datasets for semantic segmentation. B: building extraction, R: road extraction, LC: land cover mapping, and CD: change detection. The number of segments was counted on available labels.}
\label{tab:comparison}
\vspace{-5mm}
\begin{center}
\scalebox{0.85}{
\begin{tabular}{c c c c c c c c c}
\hline\hline
Image level & GSD (m) & Dataset & Task & Classes & Countries & Regions & Area ($km^2$) & Segments \\
\hline
\multirow{2}{*}{Meter level} & 10 & OpenSentinelMap~\cite{Johnson_2022_CVPR} & LC & 15 & --- & --- & 505,202 & 3,467,552 \\
& 3 & DynamicEarthNet~\cite{Toker_2022_CVPR} & LC/CD & 7 & --- & 75 & 707 & 897,855 \\
\hline
\multirow{6}{*}{Sub-meter level} & 0.3--0.5 & SpaceNet 1\&2~\cite{van2018spacenet} & B & 2 & 5 & 5 & 5,555 & 685,235 \\
& 0.5/0.3/0.5 & DeepGlobe~\cite{demir2018deepglobe} & R/B/LC & 2/2/7 & --- & --- & 2,220/984/1,717 & ---/302,701/20,697 \\
& 0.02--0.2 & Open Cities AI~\cite{OpenCitiesAI} & B & 2 & 8 & 11 & 419 & 792,484 \\
& 0.5 & xBD~\cite{gupta2019creating} & B/CD & 2/4 & 15 & 21 & 3,382 & 850,736 \\
& 0.3 & LoveDA~\cite{wang2021loveda} & LC & 7 & 1 & 3 & 536 & 166,768 \\
\cline{3-9}
& 0.25--0.5 & OpenEarthMap & LC & 8 & 44 & 97 & 799 & 2,205,395 \\
\hline\hline
\end{tabular}
}
\vspace{-7mm}
\end{center}
\end{table*}

\vspace{-0.52cm}
\section{Introduction}\label{sec:1}
Land cover classification maps are the basic information for decision making in various applications, such as land use planning, food security, resource management, and disaster response. Meter-level resolution satellite imagery have been used to map the world, as represented by GlobeLand30~\cite{chen2015global}, FROM-GLC~\cite{chen2019stable}, and recent benchmarks such as OpenSentinelMap~\cite{Johnson_2022_CVPR} and DynamicEarthNet~\cite{Toker_2022_CVPR}. Satellite imagery at a sub-meter level of ground sampling distance (GSD) enables the extraction of core map information such as buildings and roads. In recent years, there has been substantial progress in automatic construction of building footprints over large areas~\cite{sirko2021continental}.

Since the advent of deep learning~\cite{dl2021BYG}, a great deal of effort has been devoted to developing benchmark datasets for high-resolution remote sensing image analysis to facilitate advances in theory and practice. SpaceNet~\cite{van2018spacenet} and IEEE GRSS DFC~\cite{GRSS-IEEE-DFC}, among others, regularly introduce benchmark datasets to the public through competitions that drive research and development. Building detection, road detection, object detection, and land cover classification (semantic segmentation) are the most typical tasks for which these datasets are used in supervised learning~\cite{zhu:review2017,ma2019deep}. Apart from supervised learning, these datasets have been used in more realistic problems, including transfer learning~\cite{wang2021loveda}, semi-supervised learning~\cite{castillo2021semi} and weakly supervised learning~\cite{robinson2021global,li2022outcome}. Benchmark datasets that contribute to solving social problems regarding change detection and disaster damage mapping have been developed as well~\cite{fujita2017damage,gupta2019creating}.

Benchmark datasets for semantic segmentation at sub-meter level resolution have two problems: regional disparity and annotation quality. The regions included in many benchmarks are often biased toward developed countries. Thus, benchmark datasets for regions where map information is not well maintained are scarce. Two main reasons why this problem has not been easily solved are the lack of high-resolution open aerial imagery in developing countries and that commercial high-resolution satellite imagery are basically not redistributable. Other than buildings and roads, the annotation quality of land cover labeling in existing benchmarks is coarse, even though images are at sub-meter level resolution. This is due to the high cost of manually labeling sub-meter-resolution imagery in spatial detail. Thus, most of the labeling data are based on OpenStreetMap~\cite{OpenStreetMap} and open map data from local governments.

In this work, we propose OpenEarthMap, a benchmark dataset for global high-resolution land cover mapping with the goal of providing automated mapping for everyone. OpenEarthMap presents a major advance over existing data with respect to geographic diversity and annotation quality (see Table~\ref{tab:comparison}). OpenEarthMap consists of 8-class land cover labels at a 0.25--0.5m GSD of 5000 images, covering 97 regions from 44 countries across 6 continents. We adopted RGB images of some existing benchmark datasets for building detection and collected additional images for areas not covered by these benchmarks to balance the regional disparities. All images were manually labeled to ensure high-quality annotation. We evaluate the performance of state-of-the-art methods for semantic segmentation and unsupervised domain adaptation tasks and identify problem settings suitable for further technical development. In addition, lightweight models based on automated neural architectural search are investigated for cases where people requiring automated mapping have limited computational resources or for rapid mapping applications such as disaster response.

\section{The Dataset}\label{sec:2}
\subsection{Source of Imagery}\label{sec:2.1}
Our strategy is to reuse images from existing benchmark datasets as much as possible and manually annotate new land cover labels. We selected xBD~\cite{gupta2019creating}, Inria~\cite{maggiori2017can}, Open Cities AI~\cite{OpenCitiesAI}, SpaceNet~\cite{van2018spacenet}, Landcover.ai~\cite{boguszewski2021landcover}, AIRS~\cite{chen2019temporary}, GeoNRW~\cite{baier2021synthesizing}, and HTCD~\cite{shao2021sunet} datasets based on the condition that the source images are redistributable, the ground sampling distance (GSD) is equal to or less than 0.5m, and the images have geocoordinate information. If there are enough images of a region, which we defined at a scale of province or city, we sampled 50--70 images of that region at a size of 1024$\times$1024 pixels. The number of images from each dataset we adopted was determined based on the diversity and balance of the continents and countries where the images were taken. For countries and regions not covered by the existing datasets, aerial images publicly available in such countries or regions were collected to mitigate the regional gap, which is an issue in most of the existing benchmark datasets. The open data were downloaded from OpenAerialMap~\cite{OpenAerialMap} and geospatial agencies~\cite{GSI,CENEPRED}. See the supplementary for more details of attribution.

In addition to this geographic diversity, our dataset includes a mixture of images taken from different platforms, including satellite, aircraft, and UAV. For very high-resolution images with GSD less than 0.25m, we resampled the images to 0.3m or 0.5m to account for object size and visual interpretability of the captured area. Basically, the images were selected by a combination of random sampling and manual checking for each region. Moreover, if the number of images of a particular region is very large in the source benchmark dataset, we trained a segmentation model using sequentially labeled data (e.g., every 10 images) and another regression model to estimate the loss. Then, we added the images that have high values of predicted loss, as they are more difficult by a model trained with the available labels to segment.

\begin{figure}[t]
\begin{center}
   \includegraphics[width=\linewidth]{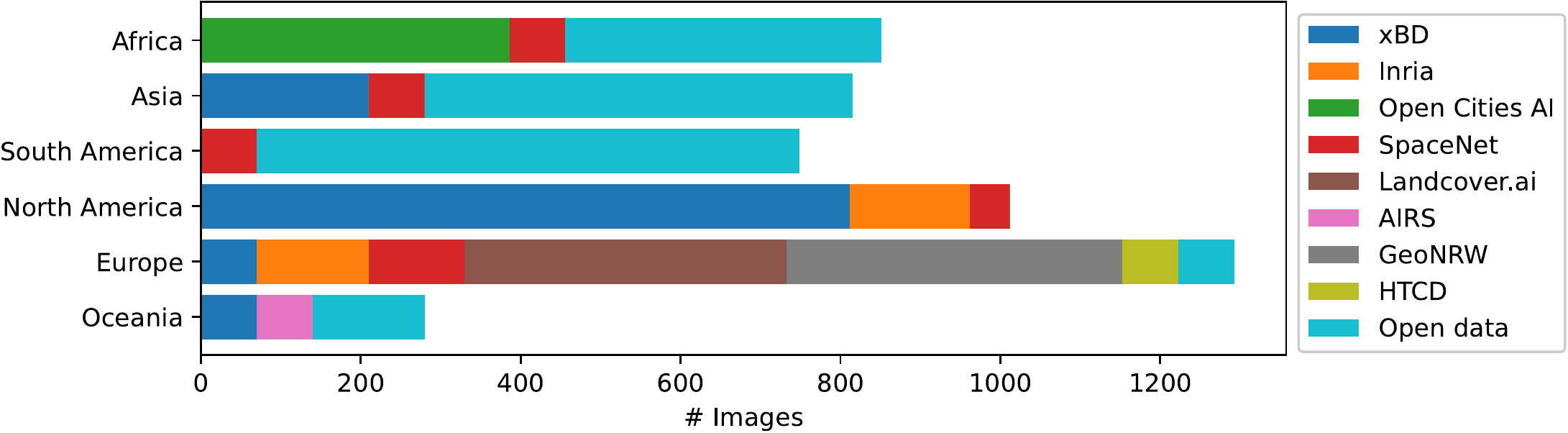}
\end{center}
\vspace{-6mm}
   \caption{The number of images of the six continents in OpenEarthMap.}
\label{fig:source}
\vspace{-1mm}
\end{figure}

\begin{figure}[t]
\begin{center}
\includegraphics[width=\linewidth]{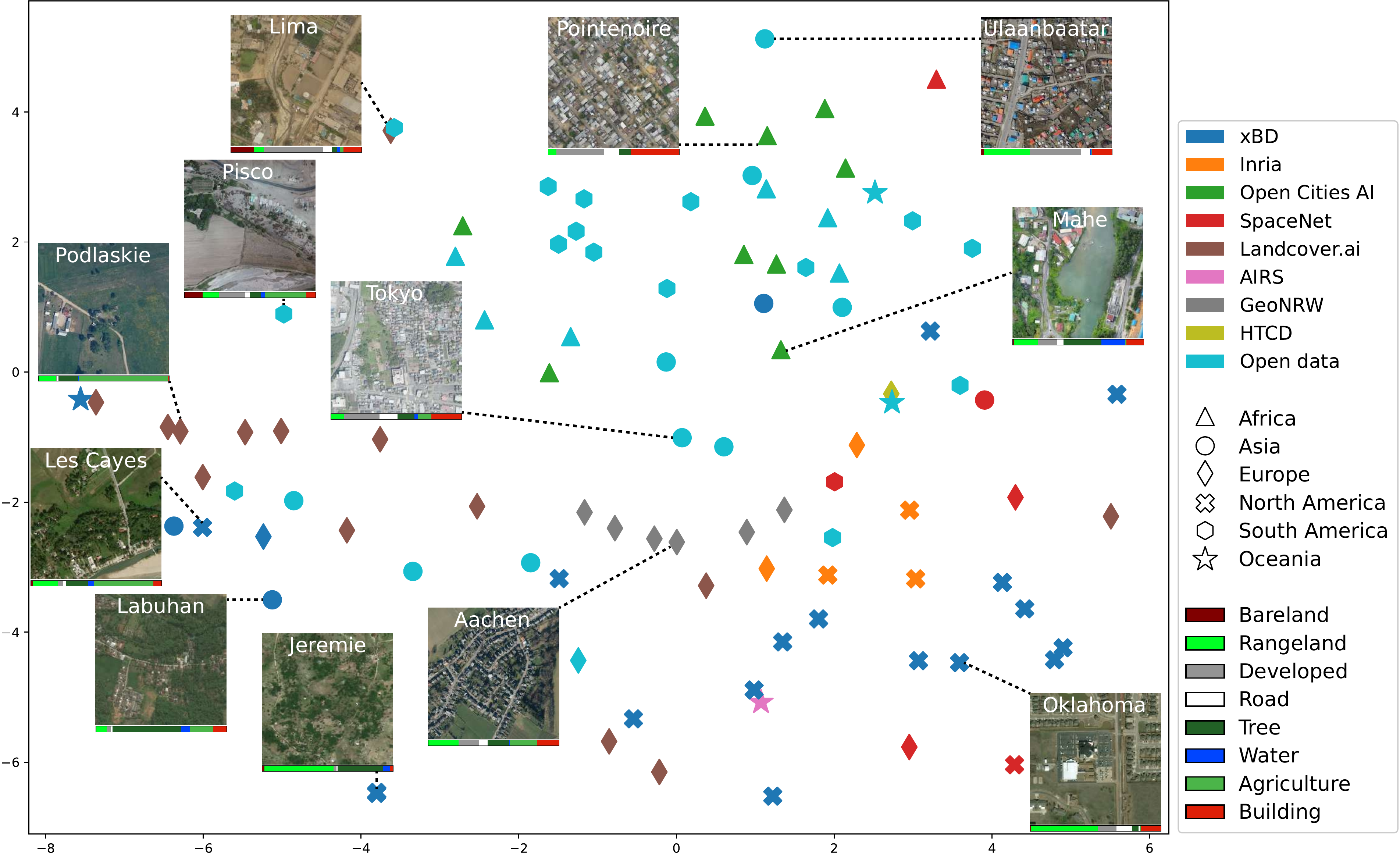}
\end{center}
\vspace{-6mm}
   \caption{t-SNE 2D visualization of the 97 regions based on features extracted with EfficientNet-B4 trained on OpenEarthMap. The images are samples of 12 regions with horizontal bar charts of class proportions attached at the bottom.}
\label{fig:tSNE_image}
\vspace{-5mm}
\end{figure}

In the end, we collected a total of 5000 images from 97 regions of six continents. Figure~\ref{fig:feature-graphic} shows annotated samples and the geographic distribution of the 97 regions with different colors indicating the source datasets. Figure~\ref{fig:source} depicts the number of images in our dataset for each of the six continents, colored to indicate the origin of the images. Asia, Africa, and South America are not well covered by the source datasets; thus, we added many images from public data to balance the regional disparities. Figure~\ref{fig:tSNE_image} presents a t-SNE 2D plot based on the similarity of image features for the 97 regions. For each region, we used the average of features extracted by EfficientNet-B4 trained as an encoder of U-Net on OpenEarthMap. The 12 representative images in the 2D plot show that different locations correspond to diverse images. It can also be seen that the different source datasets are complementary to each other, and that the diversity of images is enriched by the open data we added. The different symbols correspond to the six continents and enable the similarities between the continents to be seen. For example, regions in Europe and North America as well as Africa and South America are similar in the image features.

\subsection{Classes, Annotations, and Data Split}\label{sec:2.2}

\noindent\textbf{Classes:} We provide annotations with eight classes: \textit{bareland}, \textit{rangeland}, \textit{developed space}, \textit{road}, \textit{tree}, \textit{water}, \textit{agriculture land}, and \textit{building}. The class selection is consistent with existing products and benchmark datasets (e.g., LoveDA~\cite{wang2021loveda} and DeepGlobe~\cite{demir2018deepglobe}) with sub-meter GSD. Table~\ref{tab:oem_pixels_segments} shows the number and proportion of labeled pixels and the number of segments of each class. Here, as well as in Table~\ref{tab:comparison}, we refer to a segment as a set of connected pixels with the same label, and it was counted using OpenCV's findContours function. It can be seen that the elevated objects (e.g., \textit{tree} and \textit{building}) are finely annotated compared to the ground objects (e.g., \textit{agriculture land}). As can be seen in the horizontal bar charts of 12 representative regions in Figure~\ref{fig:tSNE_image}, the class proportions in the different regions are diverse.

\noindent\textbf{Annotations: }A total of 16 people worked on the annotation process: 8 people were responsible for annotating the images, while the remaining 8 people performed quality checks to point out errors. One person labels an image and at least two people perform the quality check. We spent a longer time labeling the first 100 images and exchanging ideas with each other to ensure that all 
participants were in agreement about the class definitions. On average, the labeling took 2.5 hours per image. This is significantly longer than the 1.5 hours of Cityscapes~\cite{cordts2016cityscapes}, which illustrates the difficulty of labelling remote sensing images.
All the labeling was done manually. For the labeling of images of the existing benchmark datasets, only the \textit{building} class was used as the starting point. However, since a lot of label noise was found, segments of \textit{buildings} were also manually modified. The most important feature of OpenEarthMap's labeling is its level of spatial detail. As shown in Table~\ref{tab:comparison}, the area covered by the images in OpenEarthMap is not very large compared to the other benchmark datasets, however, the number of segments is 10 times more  than that of LoveDA.

The accuracy of human annotations is evaluated by having two different people labeling 200 images twice. We selected two or three images with as many classes as possible based on the first annotation from each region to constitute the 200 images. The percentage of pixels that were labeled as the same class in the two different annotations by different people is 78\%. This percentage is significantly lower than the 96\% in Cityscapes~\cite{cordts2016cityscapes}, suggesting that annotation of high-resolution remote sensing images is much more challenging than annotation of urban street scenes. The relationship between human labeling accuracy and estimation accuracy of the state-of-the-art segmentation models is discussed in Section~\ref{sec:3.4}.

\definecolor{bareland}{rgb}{0.50196078,0,0}
\definecolor{rangeland}{rgb}{0,1,0.14117647}
\definecolor{develop}{rgb}{0.58039216,0.58039216,0.58039216}
\definecolor{road}{rgb}{1,1,1}
\definecolor{tree}{rgb}{0.13333333,0.38039216,0.14901961}
\definecolor{water}{rgb}{0,0.27058824,1}
\definecolor{agriculture}{rgb}{0.29411765,0.70980392,0.28627451}
\definecolor{building}{rgb}{0.87058824,0.12156863,0.02745098}

\begin{table}
\caption{The number and proportion of pixels and the number of segments of the eight classes.}
\label{tab:oem_pixels_segments}
\vspace{-5mm}
\begin{center}
\scalebox{0.7}{\begin{tabular}{c c c c c}
    \hline\hline
    Color & \multirow{2}{*}{Class} & \multicolumn{2}{c}{Pixels}& Segments\\
    \cline{3-4} (HEX) & & Count (M) & (\%) & (K) \\
    \hline
    \colorbox{bareland}{\textcolor{white}{800000}} & Bareland  & 74   & 1.5 & 6.3 \\
    \colorbox{rangeland}{\textcolor{white}{00FF24}}& Rangeland & 1130 & 22.9 & 459.4 \\
    \colorbox{develop}{\textcolor{white}{949494}}& Developed space  & 798  & 16.1 & 382.7 \\
                                FFFFFF & Road   & 331  & 6.7 & 27.9 \\
    \colorbox{tree}{\textcolor{white}{226126}} & Tree & 996  & 20.2 & 902.9\\
    \colorbox{water}{\textcolor{white}{0045FF}}& Water & 161  & 3.3 & 18.7 \\
    \colorbox{agriculture}{\textcolor{white}{4BB549}}& Agriculture land & 680  & 13.7 & 18.2 \\
    \colorbox{building}{\textcolor{white}{DE1F07}}& Building & 770  & 15.6 & 389.3 \\
    \hline\hline
    \end{tabular}}
    \vspace{-0.6cm}
\end{center}
\end{table}

\noindent\textbf{Data split: }For the semantic segmentation task, the images from each region were randomly divided into training, validation, and test sets with a ratio of 6:1:3, which respectively yielded 3000, 500, and 1500 images out of the total 5000 images. To ensure that all classes in each region are included in the training set and as many classes as possible are included in the test set, the split with the least mismatch between the training and test classes was selected from multiple random trials. For the unsupervised domain adaptation (UDA) tasks, we adopt two ways of data split to investigate regional-level and continent-wise domain gaps. For regional-level UDA, the entire dataset is divided into 73 and 24 regions for source and target domains, respectively. The split was performed in such a way that both the source and the target domains consist of relatively even distribution of the countries from all six continents as well as a balance between  urban and rural areas. This split is not as extreme as the urban-rural split in LoveDA but rather it is a realistic scenario in domain adaptation where OpenEarthMap is at hand as source data and adapts models for mapping in any new region, not only urban-rural adaptation. For continent-wise UDA, we use data from one continent as the source domain and other continents as the target domains.

\subsection{Comparison with Related Datasets}\label{sec:2.3}
\begin{figure}[t]
  \begin{minipage}[b]{0.49\linewidth}
    \centering
    \includegraphics[width=\linewidth]{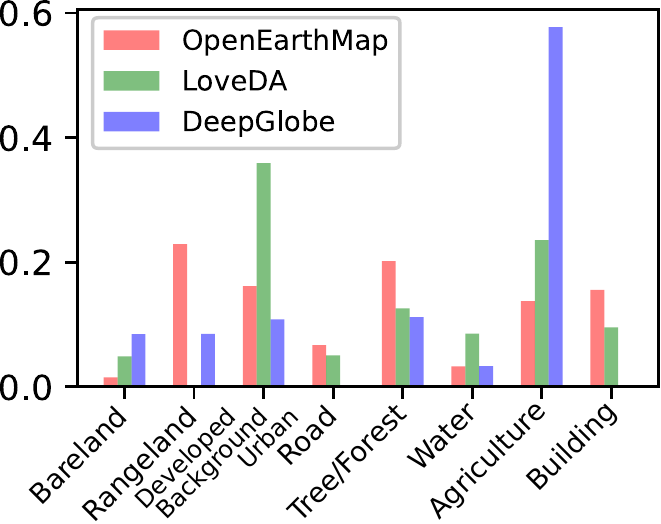}
    \subcaption{Class proportions}\label{fig:prop}
  \end{minipage}
  \begin{minipage}[b]{0.49\linewidth}
    \centering
    \includegraphics[width=\linewidth]{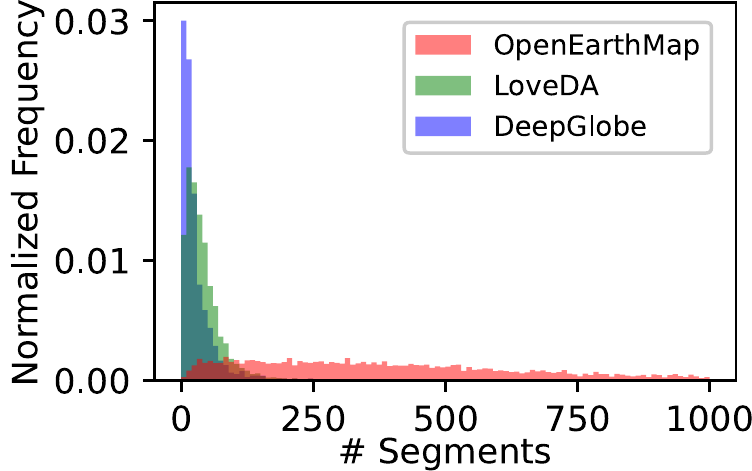}
    \subcaption{Histogram of \# segments}\label{fig:segments}
  \end{minipage}
  \vspace{-2mm}
   \caption{(a) The proportions of annotated pixels per class and (b) Normalized histograms of the number of segments for OpenEarthMap, LoveDA and DeepGlobe datasets.}
\label{fig:prop_segments}
\vspace{-0.5cm}
\end{figure}

Very recently, meter-level resolution benchmarks have made great progress in global land cover mapping; OpenSentinelMap~\cite{Johnson_2022_CVPR} is featured in its comprehensive coverage of the globe exploiting open data of Sentinel-2 and OpenStreetMap while DynamicEarthNet~\cite{Toker_2022_CVPR} is advantageous at high-temporal resolution. OpenEarthMap goes one step further in providing spatially detailed annotation at the sub-meter level. A more detailed comparison is made with LoveDA~\cite{wang2021loveda} and DeepGlobe~\cite{demir2018deepglobe}, which have similar resolution and class definitions as OpenEarthMap. Figure~\ref{fig:prop} shows a comparison of the class proportions of the three datasets. It should be noted that LoveDA does not include \textit{rangeland}, and that in the DeepGlobe dataset for land cover classification, \textit{buildings} and \textit{roads} are included in the \textit{urban} class. There is no dominant class in OpenEarthMap and the class proportions are relatively balanced. The normalized histogram of the number of segments in a single image is shown in Figure~\ref{fig:segments}. In terms of image size, LoveDA is the same (1024$\times$1024 pixels) as OpenEarthMap, while DeepGlobe is larger (2448$\times$2448 pixels). The histogram of OpenEarthMap has a very long tail, showing a much larger number of segments in each image of OpenEarthMap than the other datasets. The spatially detailed labeling of the OpeneEarthMap is reflected in the cross-dataset evaluation and the out-of-sample prediction results of trained models presented in Sections~\ref{sec:5.1} and \ref{sec:5.2}.

\section{Land Cover Semantic Segmentation}\label{sec:3}
\subsection{Baselines}\label{sec:3.1}
For the land cover semantic segmentation task, CNN-based and Transformer-based architectures were evaluated and compared on the OpenEarthMap dataset. More specifically, the chosen models are U-Net~\cite{ronneberger2015u}, U-NetFormer~\cite{WANG2022196}, FT-U-NetFormer~\cite{WANG2022196}, DeepLabV3~\cite{deeplabv3plus2018}, HRNet~\cite{SunXLW19}, SETR~\cite{SETR}, SegFormer~\cite{xie2021segformer}, and UPerNet~\cite{xiao2018unified} with backbones of ViT~\cite{dosoViTskiy2020}, Twins~\cite{chu2021twins}, Swin Transformer~\cite{liu2021Swin}, ConvNeXt~\cite{liu2022convnet}, and K-Net~\cite{zhang2021knet}. 
 
\subsection{Results}\label{sec:3.2}
\noindent\textbf{General results:} The results obtained on the test set of OpenEarthMap are presented in Table~\ref{tab:segmentation_results}. The main findings are discussed as follows: (1) U-Net with EfficientNet-B4 as backbone outperforms both U-Net with ResNet-34 and U-Net with VGG-11. The reason might be that EfficientNet-B4 is more effective for extracting relevant features, and to that effect, both high-level features and low-level spatial information are used for robust segmentation. (2) UPerNet with Swin-B and Twins, as well as SegFormer and K-Net perform better than DeepLabV3 and HRNet. This might be attributed to the strong modeling capabilities and dynamic feature aggregation of Swin-B, Twins, and MiT-B5. (3) U-NetFormer and FT-U-NetFormer share the top positions because both methods adopt a global–local Transformer block to construct global and local information in the decoder, and use advanced encoder (e.g., ResNeXt and Swin-B) to extract features. (4) UPerNet with ViT and ConvNeXt, along with SETR obtain worse results than other Transformer-based models. Two reasons might be that the hyperparameters (e.g., optimizer and learning rate) of these methods may need to be carefully tuned, and advanced data augmentation may be required for transfer learning from ImageNet to the OpenEarthMap dataset. In all, considering performance along with the number of parameters and FLOPs, U-Net-EfficientNet-B4, UPerNet-Swin-B, and FT-U-NetFormer are recommended.
\par
\begin{figure*}[t]
\vspace{-3mm}
\begin{center}
\includegraphics[width=\linewidth]{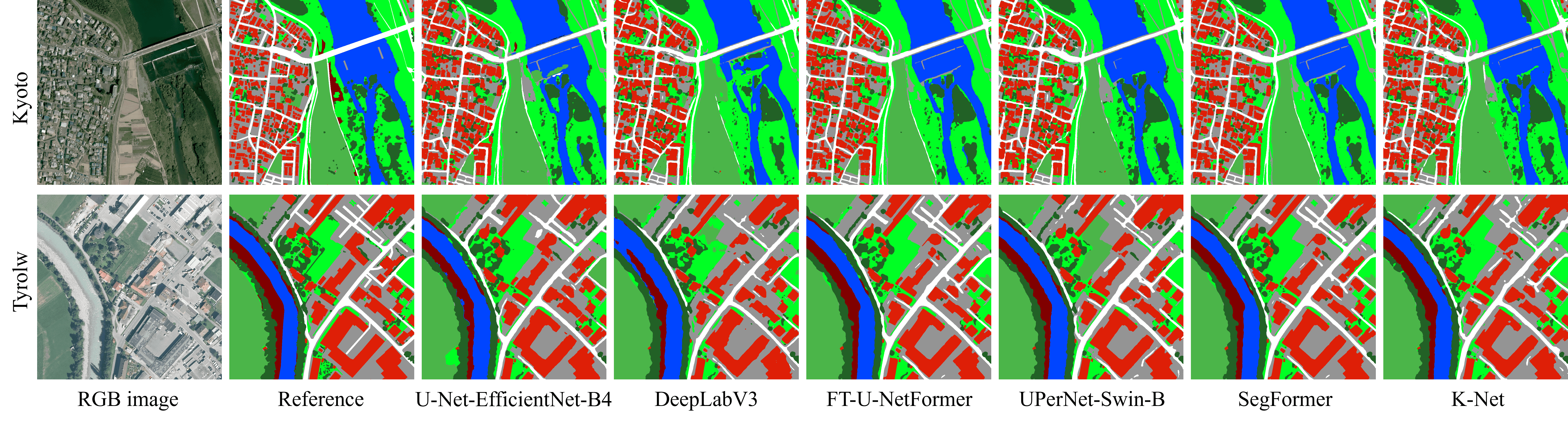}
\end{center}
\vspace{-7mm}
\caption{Visual comparison of land cover mapping results of some of the baseline models presented in Table~\ref{tab:segmentation_results}.}
\label{fig:segmentation_results}
\vspace{-3mm}
\end{figure*}

\noindent\textbf{Visualization:}
Examples of segmentation results obtained from some selected methods are presented in Figure~\ref{fig:segmentation_results}. The U-Net-EfficientNet-B4 and FT-U-NetFormer produce the best detailed visualization results. In the first row of Figure~\ref{fig:segmentation_results},  DeeplabeV3  wrongly classified the \textit{water} area of the dam as \textit{rangeland} while other methods identified them. In the second row, U-Net-EfficientNet-B4, SegFormer and FT-U-NetFormer were able to identify the tiny roads in the top-right parts of the image.
\textit{Water} and \textit{bareland} classes respectively achieved the highest and the lowest accuracies in all methods. The boundaries of the \textit{buildings} and the \textit{roads} were difficult to identify properly because of disorganized layouts and varying sizes. \textit{Rangeland}, \textit{agricultural land} and \textit{trees} are easy to confuse due to the similarities in their spectra. \textit{Roads} were easily misclassified as \textit{developed space} because parking lots and cover materials in some rural areas are quite similar.
\par

\addtolength{\tabcolsep}{-2pt}
\begin{table*}
\captionsetup{width=.9\linewidth}
\caption{Semantic segmentation results of the baseline models on the test set of the OpenEarthMap dataset. The results are based on test-time augmentation (TTA), in particular flipping.}
\label{tab:segmentation_results}
\vspace{-5mm}
\begin{center}
\scalebox{0.75}{
\begin{tabular}{c c c c c c c c c c c c c}
\hline \hline
	\multirow{2}{*}{Method}& \multirow{2}{*}{Backbone} &	 \multicolumn{8}{c}{IoU (\%)}  & {mIoU} & {Params} & {FLOPs}
	\\ \cline{3-10}
	& &	Bareland	&	Rangeland	&	Developed	&	Road	&	Tree	&	Water	&	Agriculture	&	Building	& (\%) & (M) & (G) \\	\hline
U-Net & VGG-11          &  40.69 &  56.76 &  53.99 &  62.16 &  72.44 &  82.81 &  73.14 &  77.77  &  64.97   &   \textbf{18.26}   &   233.33  \\
U-Net & ResNet-34       &  40.35 &  57.75 &  54.92 &  62.87 &  72.65 &  82.24 &  74.06 &  78.58  &  65.43   &   24.44   &   126.68  \\
U-Net & EfficientNet-B4 &  50.63 &  58.17 &  56.27 &  64.83 &  73.20 &  86.02 &  76.28 &  80.20  &  68.20   &   20.30   &   \textbf{45.47}   \\		
U-NetFormer & ResNeXt101&   46.09&	60.67&	\textbf{58.12}&	65.07&	\textbf{73.77}&	86.34&	76.98&	79.96&	68.37& 192.71  &   769.25  \\
FT-U-NetFormer & Swin-B &  \textbf{50.19}&	\textbf{60.84}&	57.58&	\textbf{65.85}&	73.33&	\textbf{87.44}&	\textbf{77.50}&	\textbf{80.29}&	\textbf{69.13}&  95.98   &   498.37  \\
DeepLabV3& ResNet-50	&	39.11&	56.16&	52.28&	60.57&	71.25&	79.32&	70.75&	75.83&	63.16	&	68.14	&	269.76	\\
HRNet & W48	            &	39.71&	55.50&	53.49&	59.22&	71.10&	79.03&	71.38&	75.12&	63.07&	65.89	&	94.06	\\
UPerNet & ViT	        &	34.39&	54.45&	50.64&	54.57&	69.73&	79.24&	66.22&	74.92&	60.52&
	144.17	&	395.07	\\
UPerNet & Swin-B	    &	44.52&	58.98&	54.78&	63.43&	72.20&	83.71&	72.97&	78.11&	66.09&
	59.94	&	236.08	\\
SegFormer	& MiT-B5    &	36.84&	57.94&	53.53&	63.60&	70.51&	80.11&	72.21&	77.35&	64.01&	81.97	&	51.86	\\
SETR PUP &ViT-L	        &	45.35&	55.72&	51.31&	55.47&	67.63&	73.12&	67.14&	75.48&	61.40&
	309.35	&	212.45	\\
UPerNet & Twins	        &	37.29&	57.62&	53.83&	60.23&	72.32&	81.93&	71.71&	77.49&	64.05&
	90.96	&	250.91	\\
UPerNet & ConvNeXt	    &	40.61&	54.94&	51.76&	58.47&	70.44&	75.95&	68.94&	74.30&	61.93&
	122.1	&	292.42	\\
K-Net	& Swin-B        &	44.02&	57.81&	54.85&	62.91&	71.76&	85.18&	73.41&	78.91&	66.11	&	246.97	&	419.51	\\

\hline \hline
\end{tabular}
}
\vspace{-1mm}
\end{center}
\end{table*}

\begin{table}
	\caption{Lightweight models discovered on OpenEarthMap training set. FLOPs and FPS are measured on 1024$\times$1024 input, and mIoU on the test set of OpenEarthMap.}
	\label{tab:nas_results}
	\vspace{-5mm}
	\setlength{\tabcolsep}{4.5pt}
	\begin{center}
	\scalebox{0.85}{
		\begin{tabular}{ccccccc}
			\hline\hline
			\multirow{2}{*}{Method} & \multirow{2}{*}{Trial} & Params & FLOPs & FPS & \multicolumn{2}{c}{mIoU (\%)} \\
			\cline{5-6}
			& & (M) & (G) & (ms) & No TTA & TTA \\
			\hline
    		\multirow{2}{*}{SparseMask} & 1st & 2.96 & \textbf{10.28} & 51.2 & 58.23 & \textbf{60.21} \\
			& 2nd & 3.10 & 10.39 & 52.2 & 58.06 & 60.00\\
			\hline
			\multirow{2}{*}{FasterSeg} & 1st & \textbf{2.23} & 14.58 & 143.2 & 57.55 & 58.35\\
			 & 2nd & 3.47 & 15.37 & \textbf{171.3} & \textbf{58.51} & 59.41\\
			\hline\hline
		\end{tabular}}
		\vspace{-9mm}
	\end{center}
\end{table}

\subsection{Neural Architecture Search}\label{sec:3.3}
LoveDA~\cite{wang2021loveda}, DeepGlobe~\cite{demir2018deepglobe}, and other previous benchmarks~\cite{boguszewski2021landcover,maggiori2017can,chen2019temporary} were experimented with only manually designed networks~\cite{ronneberger2015u,SunXLW19,deeplabv3plus2018,longFCN2015,lin2017feature} for the semantic segmentation task. In contrast, we further experimented the OpenEarthMap dataset with two automated neural architecture search methods, SparseMask~\cite{Wu2019SparseMask} and FasterSeg~\cite{Chen2020FasterSeg}, by automatically searching for compact segmentation architectures. Such architectures might offer a useful baseline for research in the field of automated neural architecture search in remote sensing with OpenEarthMap.
Following the architecture search protocols in both methods (see the supplementary for more details), we searched for lightweight segmentation networks on the OpenEarthMap dataset. Four experiments were performed, two with each method, and the results are presented in Table~\ref{tab:nas_results}. Both methods were able to discover compact networks, however, FasterSeg discovered the lightest-weight network. The networks discovered by SparseMask have less computational complexity but with low inference speed. Whereas FasterSeg networks have high computation cost and high inference speed. For real-time mapping (no TTA), FasterSeg might serve as a baseline for the OpenEarthMap dataset. For non real-time mapping (where TTA is used), SparseMask might be adopted as a baseline. Compared to the manually designed baseline models presented in Table~\ref{tab:segmentation_results}, the lightweight discovered networks ($<4M$ params) competed with UPerNet-ViT ($144.17M$ params) and trailed behind FT-U-NetFormer ($95.98M$ params) by approximately 9\% accuracy rate. 
 
\subsection{Human Annotation \textit{vs} Machine Prediction}\label{sec:3.4}
\begin{figure}[t]
  \begin{minipage}[b]{0.49\linewidth}
    \centering
    \includegraphics[width=\linewidth]{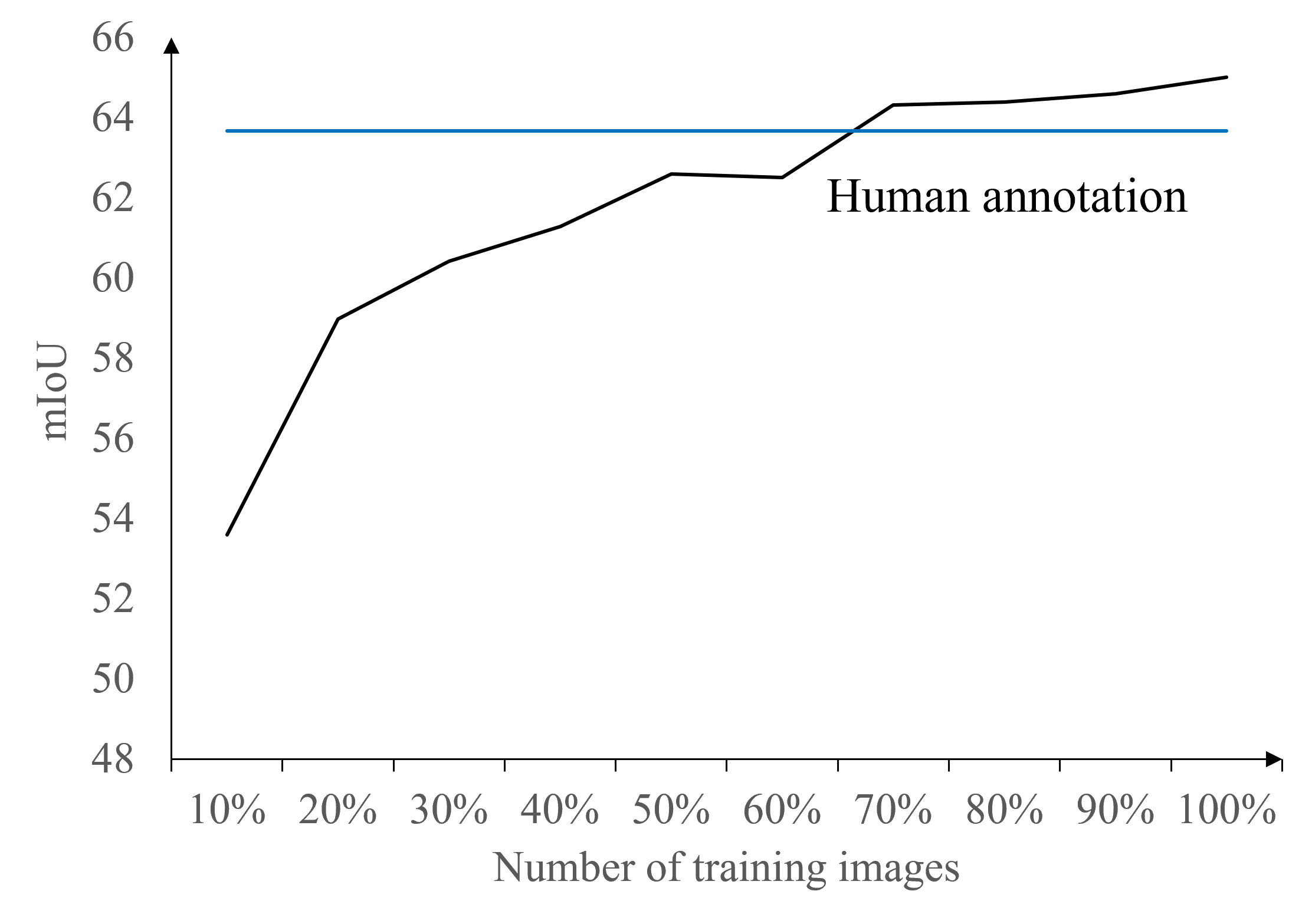}
    \vspace{-6mm}
    \subcaption{mIoU}\label{fig:human_mIoU}
  \end{minipage}
  \begin{minipage}[b]{0.49\linewidth}
    \centering
    \includegraphics[width=\linewidth]{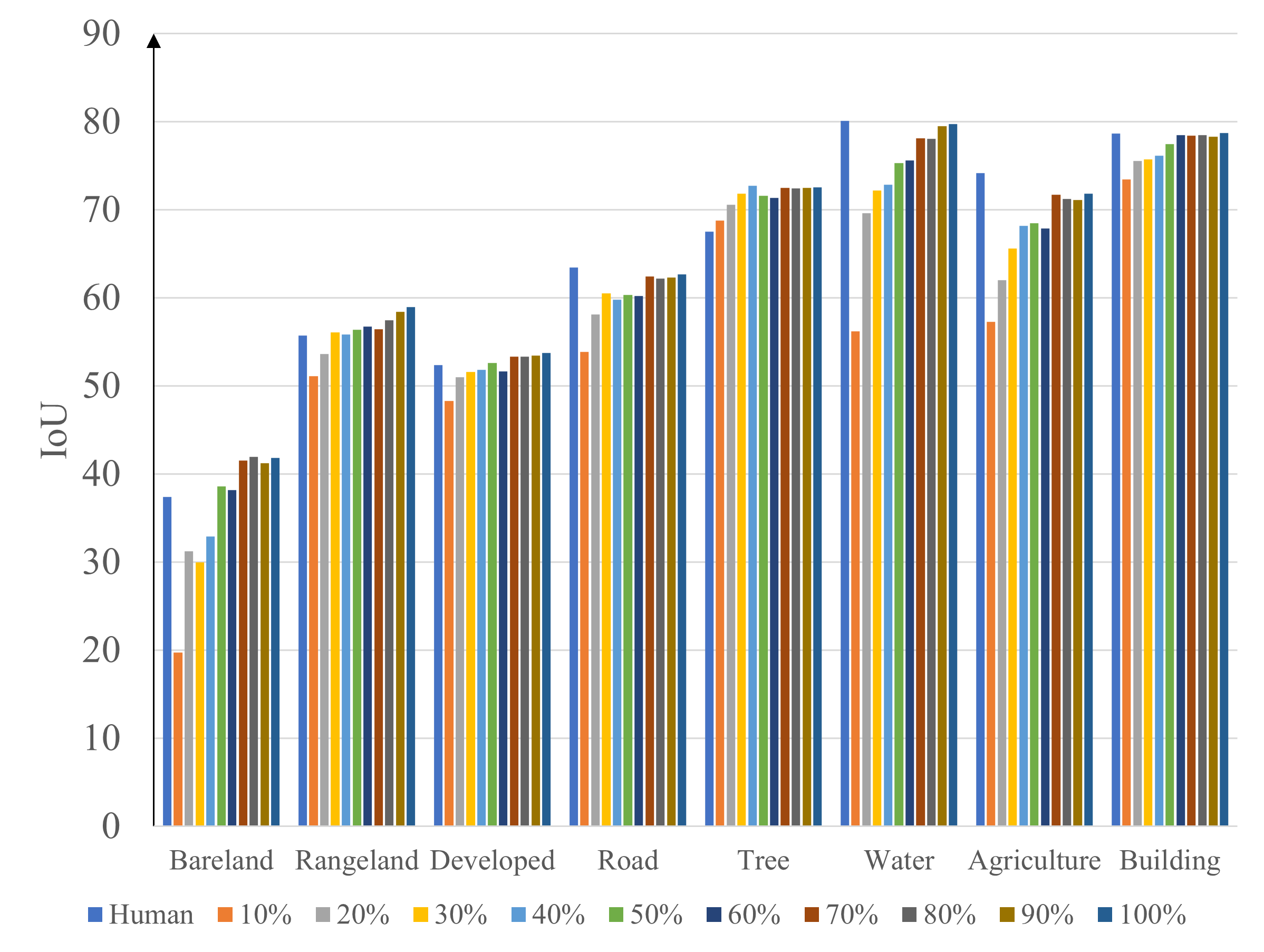}
    \vspace{-6mm}
    \subcaption{Class-specific IoUs}\label{fig:human_ious}
  \end{minipage}
  \vspace{-2mm}
   \caption{Human annotation \textit{vs} machine predictions with varying numbers of images from the training set. Human annotation accuracies mean the IoUs between two different human annotations.}
\label{fig:hum_results}
\vspace{-5mm}
\end{figure}

As mentioned in Section~\ref{sec:2.2}, 200 images were labeled twice by different people. The remaining 4800 images were used to train UPerNet with Swin-B to compare the quality of human labeling with the results from the machine. To effectively investigate the comparison,  the number of training images was varied from 10\% to 100\%; the results are shown in Figure~\ref{fig:hum_results}. It can be seen that with 50\% of the training images, the machine attains almost the same level of human annotation and larger training percentages improve the accuracy (see Figure~\ref{fig:human_mIoU}). For human annotation, the challenging classes include \textit{bareland}, \textit{rangeland}, and \textit{tree}. For \textit{bareland}, \textit{rangeland}, \textit{developed space}, and \textit{tree} classes, 50\%, 30\%, 50\%, and 10\% of the training set, respectively, yielded better results than the ones of human annotation (see Figure~\ref{fig:human_ious}). The challenging class for the machine is \textit{agriculture land}, where it trails behind human annotation by 2.3\%. Regarding \textit{road}, \textit{water}, and \textit{building} classes, with 100\% of the training images, the machine slightly ($<0.34\%$) trails behind the human annotation.

\subsection{Learning from Limited Labels}
We also investigated the performance of CNN-based (U-Net-EfficientNet-B4) and Transformer-based (SegFormer, UPerNet-Swin-B and K-Net) models on limited training samples. Table~\ref{tab:segmentation_results_10} presents the results of using only 10\% of the OpenEarthMap training set to train the models. It is apparent from Table~\ref{tab:segmentation_results_10} that U-Net-EfficientNet-B4 outperforms all three Vision Transformer-based methods in all the class-specific IoUs by about 6-15\%.The main reason is that the representation capacity of ViTs typically lacks the inductive bias in CNNs. Therefore, ViTs require more training data than CNNs~\cite{liu2021efficient,dosoViTskiy2020}. We believe that Vision Transformers with small-sized data~\cite{liu2021efficient} or limited labels~\cite{NEURIPS2021_9a49a25d} is an interesting topic that requires further study. Moreover, data augmentation, regularization, and tuning of hyper-parameters still need to be explored when training on limited training data~\cite{steinerhow}.

\section{Unsupervised Domain Adaptation}\label{sec:4}

\subsection{Baselines}\label{sec:4.2}
For the unsupervised domain adaptation task, a metric-based method (MCD~\cite{TzengHZSD14}), adversarial training methods including AdaptSeg~\cite{Tsai_adaptseg_2018}, category-level adversarial network (CLAN)~\cite{luo2019Taking}, TransNorm~\cite{Wang19TransNorm}, and fine-grained adversarial learning framework for domain adaptive (FADA)~\cite{Haoran_2020_ECCV}), as well as self-training methods including pyramid curriculum DA (PyCDA)~\cite{Lian_2019_ICCV}, class-balanced self-training (CBST)~\cite{zou2018unsupervised}, instance adaptive self-training (IAST)~\cite{mei2020instance}, and DAFormer~\cite{hoyer2022daformer} are adopted. DAFormer is based on SegFormer and the others are based on DeepLabV2. \par

\subsection{Results}\label{sec:4.3}
\begin{table}
\caption{Semantic segmentation results of selected baseline models trained on only 10\% of OpenEarthMap training set.}
\label{tab:segmentation_results_10}
\vspace{-5mm}
\addtolength{\tabcolsep}{-3pt}
\scriptsize
\begin{center}
\begin{tabular}{c c c c c c c c c c}
	\hline\hline
     & Bare	&	Range	&	Dev	&	Road	&	Tree	&	Water	&	Agri	&	Building	& mIoU\\	
     \hline
	U-Net-EfficientNet-B4 &	\textbf{32.62} & \textbf{52.43} & \textbf{49.77} & \textbf{58.47} & \textbf{69.26} & \textbf{74.39} & \textbf{70.16} & \textbf{74.35} & \textbf{60.18} \\ 
	SegFormer &	16.15	&	44.08	&	45.88	&	51.39	&	65.72	&	61.42	&	58.54	&	69.71	&	51.61	\\
	UPerNet-Swin-B   & 18.32	&	47.82	&	48.2	&	53.46	&	66.89	&	59.62	&	55.22	&	69.55	&	52.39	\\
	K-Net &	18.62	&	50.26	&	48.93	&	55.22	&	66.45	&	60.76	&	62.06	&	72.33	&	54.33	\\ 
	\hline \hline
\end{tabular}
\vspace{-5mm}
\end{center}
\end{table}
\addtolength{\tabcolsep}{0pt}

\begin{table}
\caption{Unsupervised domain adaptation results obtained on the test set of 24 regions in the OpenEarthMap dataset.}
\label{tab:uda_results}
\vspace{-5mm}
\begin{center}
\scalebox{0.62}{
\begin{tabular}{c c c c c c c c c c c}
    \hline \hline
	& \multirow{2}{*}{Type}	& \multicolumn{8}{c}{IoU (\%)} & \multirow{2}{*}{mIoU (\%)}	\\ 
	  \cline{3-10}
	&	&	Bare	&	Range	&	Dev	&	Road	&	Tree	&	Water	&	Agri	&	Build & \\
	\hline
    \multicolumn{11}{c}{DeeplabV2-based} \\ \hline		
    Oracle	&	—	&	37.06&	43.65&	38.03&	43.12&	61.61&	73.89&	75.90&	63.93&	54.65\\
    Source only 	&	—	&	26.86&	42.14&	36.48&	42.03&	58.58&	61.35&	70.77&	61.87&	50.01\\
    MCD	&	—	&	16.77&	41.55&	35.89&	44.24&	56.15&	57.84&	62.57&	63.83&	47.36\\
    AdaptSeg &	AT	&	28.77&	41.47&	36.09&	45.16&	46.65&	34.48&	68.47&	63.74&	45.60\\
    FADA	&	AT	&	26.29&	37.91&	34.91 &	37.13&	54.19&	40.68&	65.36&	58.32&	44.35\\
    CLAN	&	AT	&	22.90&	42.25&	39.49&	44.12&	58.98&	58.99&	59.51&	64.53&	48.85\\
    TransNorm	&	AT	&	27.54&	45.13&	37.99&	45.56&	57.06&	63.84&	66.26&	64.71&	51.01\\
    PyCDA	&	ST	&	21.95&	32.33&	22.89 &	34.81&	44.95&	34.16&	56.74&	55.31&	37.89 \\
    CBST	&	ST	&	29.64&	43.79&	37.99&	49.19&	57.33&	60.75&	71.93&	65.46&	52.01 \\
    IAST	&	ST	&	33.68&	43.64&	37.03&	45.16&	59.61&	72.08&	74.72&	61.77&	53.46\\ 
    \hline
    \multicolumn{11}{c}{SegFormer-based} \\ 
    \hline
    Oracle	&  — &	 43.14  & 53.02 &	51.50&	61.13&	68.06&	81.89&	81.38&	79.81&	64.99 \\
    Source only & — &	28.37&	48.96&	46.49&	54.05&	67.62&	75.32&	77.93&	75.79&	59.32 \\
    DAFormer	&	ST	&	\textbf{37.16}&	\textbf{51.07}&	\textbf{50.36}&	\textbf{58.07}&	\textbf{68.3}4&	\textbf{78.39}&\textbf{78.08}&	\textbf{77.30}&	\textbf{62.35} \\
    \hline \hline
\end{tabular}}
\vspace{-8mm}
\end{center}
\end{table}

\begin{figure*}[!htt]
\begin{center}
\includegraphics[width=\linewidth]{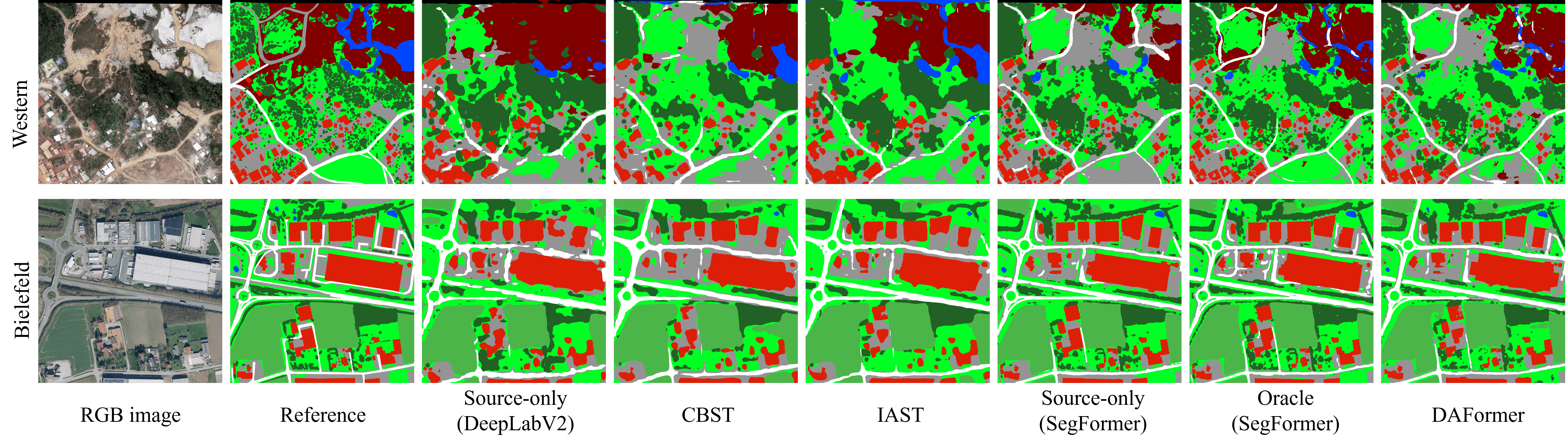}
\end{center}
\vspace{-7mm}
\caption{Visual comparison of unsupervised domain adaption results of some of the baseline models presented in Table~\ref{tab:uda_results}.}
\label{fig:uda_results}
\vspace{-2mm}
\end{figure*}

\noindent\textbf{Regional-level UDA}: We investigated the regional-level domain gap since different regions in the same continent might suffer from a distribution shift. The results obtained on the test set of 24 regions of OpenEarthMap are presented in Table~\ref{tab:uda_results}. In general, the Oracle settings obtained the best results. Due to the regional domain gap, the source-only settings yielded the lowest accuracy. The results of source-only SegFormer are significantly better than those of source-only DeepLabV2. Compared to manufactured classes (i.e., \textit{building} and \textit{road}), the accuracies of natural classes (i.e., \textit{water} and \textit{bareland}) decreased significantly. With the exception of TransNorm, the adversarial training methods did not perform well on this task due to the diversity in the OpenEarthMap dataset. TransNorm slightly improved the performance because the source and the target images have distinct spectral statistics since they were taken from different sensors and regions. The class imbalance problem is addressed using pseudo-label creation via the CBST and IAST techniques, resulting in higher performance. Due to better domain generalization of SegFormer and effective training strategy in self-training, DAFormer obtained the best mIoU of 62.35\%. Visual examples of the UDA results are presented in Figure~\ref{fig:uda_results}. 
In the first row of Figure~\ref{fig:uda_results}, source-only DeepLabV2 can barely identify the \textit{water} area (top-right) and the roads (bottom-right). IAST and CBST performance improves for \textit{water} but they lose the ability to recognize the roads. DAFormer performs very well in the two complex areas. In the second row, DAFormer shows better visualization results in the small \textit{water} area (top-right) and the boundaries of \textit{roads} and \textit{buildings} than the other UDA methods.
\par

\noindent\textbf{Continent-wise UDA:} 
\begin{figure}
\begin{center}
\includegraphics[width=\linewidth]{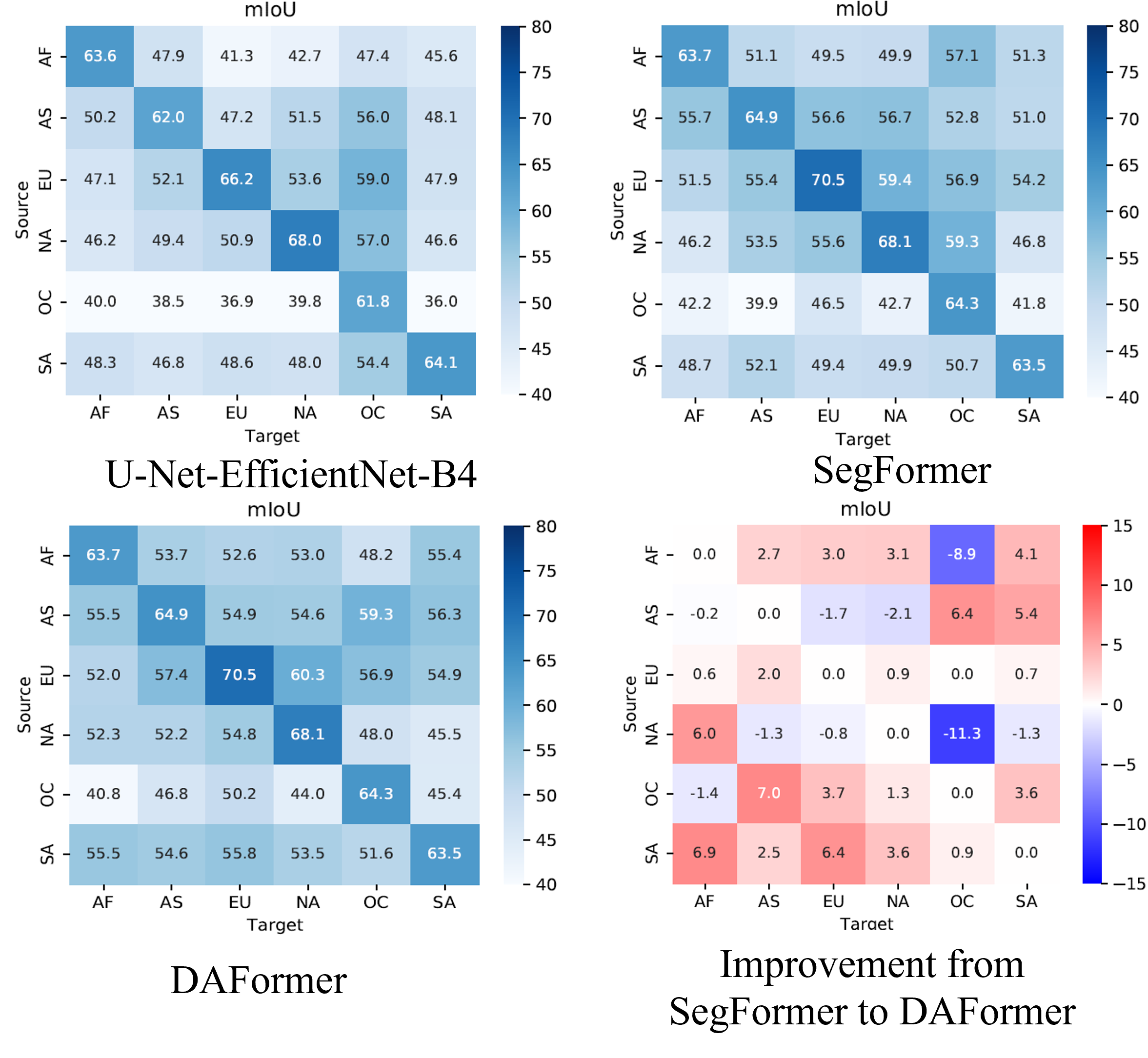}
\end{center}
\vspace{-7mm}
\caption{Continent-wise UDA results. Asia: AS, Europe: EU, Africa: AF, North America: NA, South America: SA, and Oceania: OC.}
\label{fig:UDA_results_continet}
\vspace{-7mm}
\end{figure}
We also investigated the continent-wise domain gap on the OpenEarthMap dataset using U-Net-EfficientNet-B4, SegFormer, and DAFormer. The results are presented in Figure~\ref{fig:UDA_results_continet}. Compared to the UDA settings (e.g., GTA5$\rightarrow$Cityscapes with similar content and different style) in computer vision and previous settings in remote sensing (e.g., urban$\rightarrow$rural in LoveDA), UDA on continent-wise has larger content and style gaps. The limited data of Oceania (OC) led to the lowest transferred results when OC is treated as the source domain. In contrast, the performance with OC as the target domain is better than other settings. Expect OC, U-Net-EfficientNet-B4 and SegFormer indicated two minor domain gaps: Europe (EU)-to-North America (NA) and Asia (AS)-to-NA. The most prominent domain gap revealed by EfficientNet-B4 and SegFormer is Africa (AF)-to-EU and NA-to-AF, respectively. For challenging UDA settings, SegFormer is generally better than U-Net-EfficientNet-B4 (26 out of 30), which is the opposite of the results in the semantic segmentation (see Table~\ref{tab:segmentation_results}) and the regional UDA setting (see Table~\ref{tab:uda_results}). DAFormer improved the results compared to SegFormer in many cases (20 out of 30). DAFormer on AF-to-OC and NA-to-OC achieved significantly poor results due to the collapsed construction of pseudo labels in the limited data of OC. Thus, challenging continent-wise UDA settings are worth exploring and possible solutions may include the extension of DAFormer or new UDA method with U-Net-EfficientNet-B4.

\section{Cross-Dataset Evaluation}\label{sec:5.1}
\begin{figure}[t]
  \begin{minipage}[b]{0.49\linewidth}
    \centering
    \includegraphics[width=\linewidth]{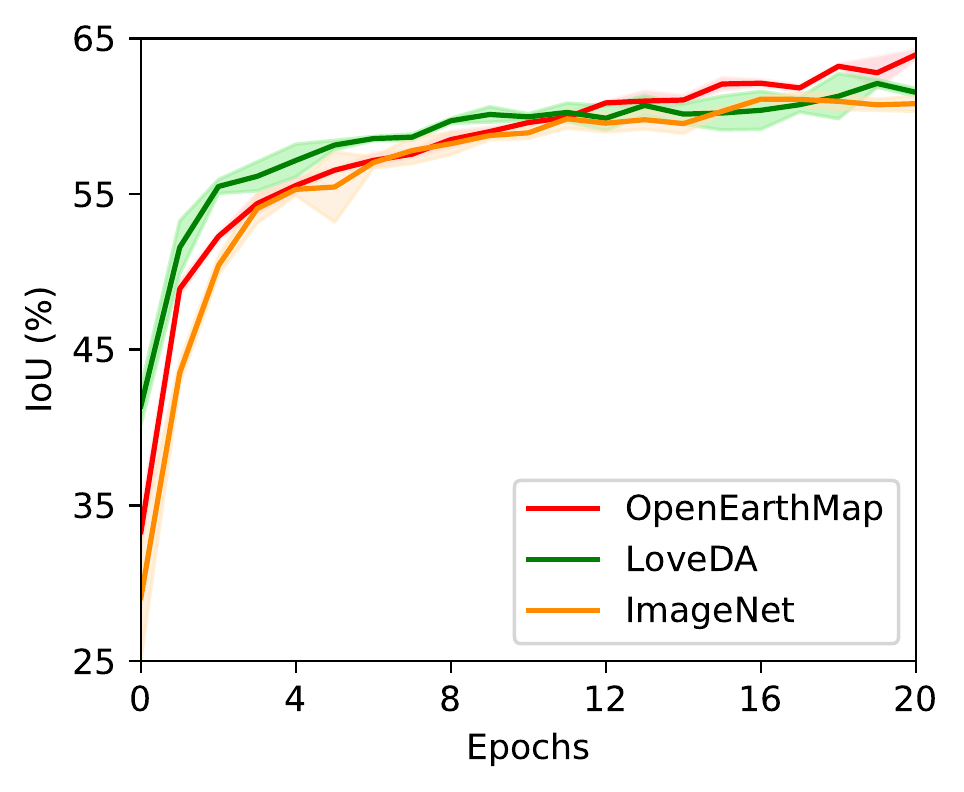}
    \vspace{-6mm}
    \subcaption{Fine-tuning on DeepGlobe}\label{fig:tune_deepglobe}
  \end{minipage}
  \begin{minipage}[b]{0.49\linewidth}
    \centering
    \includegraphics[width=\linewidth]{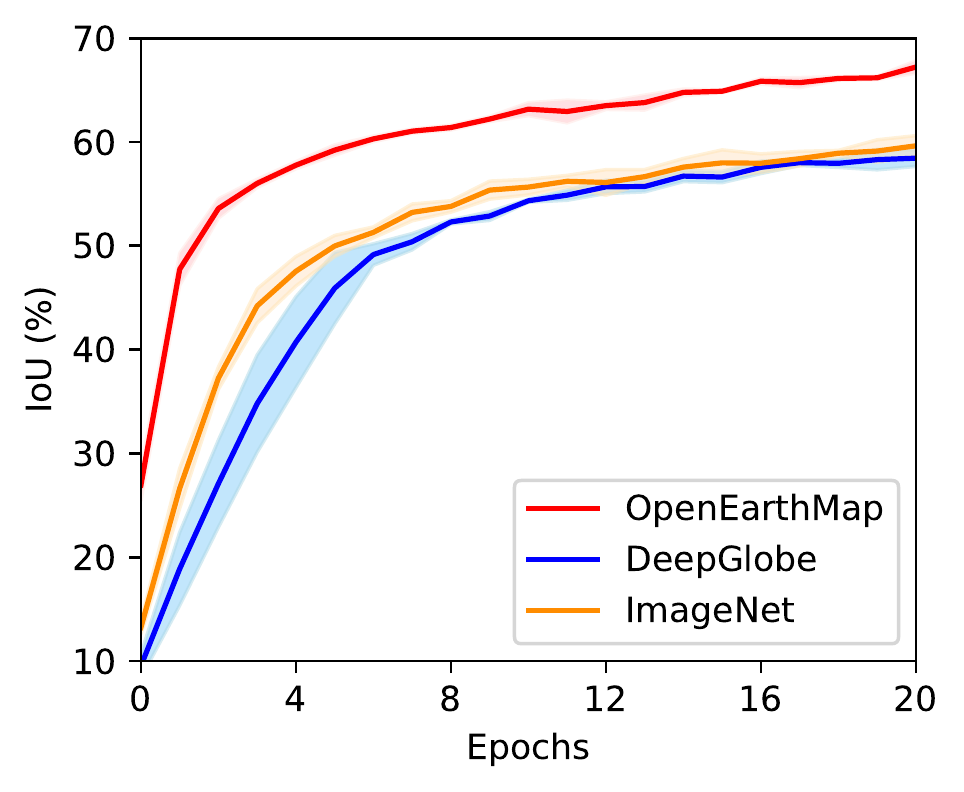}
    \vspace{-6mm}
    \subcaption{Fine-tuning on LoveDA}\label{fig:tune_loveda}
  \end{minipage}
  \vspace{-2mm}
 \caption{Comparison among OpenEarthMap, LoveDA and DeepGlobe pre-trained models.}
\label{fig:crossdataset}
\vspace{-4mm}
\end{figure}

In this section, we evaluate the advantage of using the OpenEarthMap dataset as a starting point (fine-tuning) in the semantic segmentation task over other open-source land cover mapping datasets. Here we compare OpenEarthMap with LoveDA~\cite{wang2021loveda} and DeepGlobe~\cite{demir2018deepglobe}. We adopted the same U-Net model with an EfficientNet-B4 as a backbone listed in Table~\ref{tab:segmentation_results} and trained from scratch on the three datasets using similar training settings as Section~\ref{sec:3.2}. Then, we fine-tuned the OpenEarthMap and the LoveDA pre-trained models on the DeepGlobe dataset. Similarly, the OpenEarthMap and the DeepGlobe pre-trained models were fine-tuned on the LoveDA dataset. All the experiments were run threefold; we report the mean and the standard deviation segmentation accuracy for 20 epochs. As presented in Figure~\ref{fig:crossdataset}, the results indicate that using a model that is pre-trained on OpenEarthMap as a starting point could yield better performance than models pre-trained on LoveDA and DeepGlobe. For example, when fine-tuned on the DeepGlobe dataset, the initial IoU score of the OpenEarthMap pre-trained model is about 4\% higher  than the fully-trained model on DeepGloble (see Figure~\ref{fig:tune_deepglobe}). Although the OpenEarthMap pre-trained model is slightly lower than the LoveDA pre-trained one in early epochs, OpenEarthMap increasingly outperforms both models as the number of epochs increases. Furthermore, when fine-tuned on the LoveDA dataset, the OpenEarthMap pre-trained model attains a more than 20\% increase in the initial IoU score, and its performance remains higher when the number of epochs increases (see Figure~\ref{fig:tune_loveda}).

\section{Demonstration on Out-of-Sample Imagery}\label{sec:5.2}
\begin{figure}[t]
\begin{center}
\includegraphics[width=\linewidth]{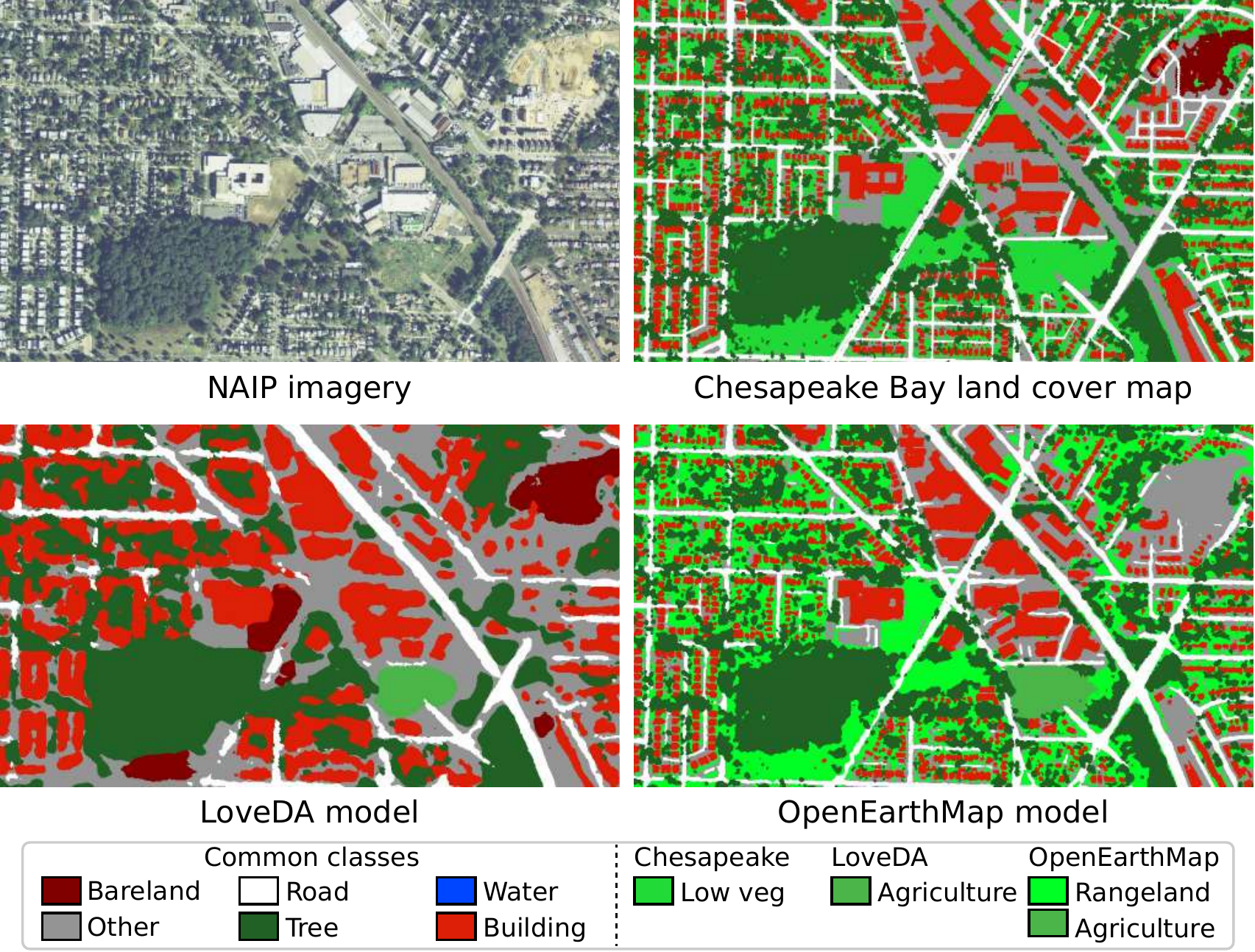}
\end{center}
\vspace{-5mm}
   \caption{Visual comparison of Chesapeake Bay land cover map with land cover maps generated by U-Net models trained on LoveDA and OpenEarthMap. The NAIP image is the source data.}
\label{fig:chesapeake}
\vspace{-5mm}
\end{figure}

To further investigate the generalization performance of a model trained on OpenEarthMap, we created land cover classification maps from out-of-sample imagery (i.e., images that are not included in OpenEarthMap). See the supplementary for more results. Here we present a map created from an NAIP~\cite{NAIP} image resampled at 0.5m GSD. A Chesapeake Bay land cover map~\cite{ChesapeakeBay} was used as reference to evaluate performance. The Chesapeake Bay land cover map consists of 13 classes. To fairly compare the mapping results of the Chesapeake Bay land cover tool with those produced by U-Net models trained on OpenEarthMap and LoveDA, we adopted six common classes (\textit{bareland}, \textit{other}, \textit{road}, \textit{tree}, \textit{water}, and \textit{building}) among the datasets and performed quantitative evaluation. Table~\ref{tab:chesapeake} shows the accuracy of the land cover mapping for an area of approximately $15 km \times 28 km$ in US, spanning from Washington, DC to Maryland. The IoUs from OpenEarthMap model are significantly higher than those from LoveDA, and the scores are high enough for practical mapping except \textit{bareland}. The accuracy of \textit{bareland} is low due to inconsistency in class definitions. For example, in the Chesapeake Bay land cover map, a construction site is labeled as \textit{bareland}, while OpenEarthMap labels the same area as \textit{developed space}. Figure~\ref{fig:chesapeake} shows a visual example of the mapping results. Note that unlike the quantitative evaluation in Table~\ref{tab:chesapeake}, the vegetation classes (\textit{low vegetation}, \textit{agriculture land}, and \textit{rangeland}) that differ among the datasets are visualized in different colors. The OpenEarthMap model result is similar to the Chesapeake Bay land cover map in both classification and resolution, and achieved very fine spatial segmentation compared to the LoveDA model. This demonstrates the advantage of OpenEarthMap over LoveDA and how finely OpenEarthMap's annotations are spatially detailed.

\begin{table}
\caption{Generalization performance of models trained on OpenEarthMap and LoveDA, and evaluated with IoU (\%) using the Chesapeake Bay high-resolution land cover map.}
\label{tab:chesapeake}
\vspace{-5mm}
\begin{center}
\scalebox{0.8}{
\begin{tabular}{c c c c c c c c}
\hline\hline
Dataset & Bare & Other & Road & Tree & Water & Build & mIoU \\
\hline
OpenEarthMap & \textbf{9.29} & \textbf{58.27} & \textbf{49.29} & \textbf{75.72} & \textbf{85.46} & \textbf{63.44} & \textbf{56.91} \\
LoveDA       & 3.07 & 40.14 & 37.71 & 69.34 & 80.12 & 45.85 & 46.04 \\
\hline\hline
\end{tabular}
}
\vspace{-8mm}
\end{center}
\end{table}

\section{Conclusion and Societal Impacts}\label{sec:6}
The existing benchmarks for land cover classification at sub-meter resolution lack regional diversity and annotation quality. To address this problem, we introduce OpenEarthMap, a benchmark dataset, for global high-resolution land cover mapping.
The diversity of the dataset is shown in the coverage of 97 regions from 44 countries across 6 continents, while its finely detailed annotations are reflected in the generalization of the feature space. 
To demonstrate the practical usefulness of OpenEarthMap, we perform baseline experiments with several state-of-the-art models for semantic segmentation and UDA tasks, and create land cover maps for out-of-sample imagery to show that models trained on OpenEarthMap can adapt and generalize across the globe. We also demonstrate the challenges of the continent-wise domain gap and limited data training. We experiment NAS-based lightweight models for mapping with resource-limited devices. Further technical development is needed to improve the performance in continent-wise domain adaptation, limited training data, and lightweight models on OpenEarthMap for worldwide evaluation. The dataset will be made publicly available for other researchers to build on it and create new practical tasks.

\noindent\textbf{Societal Impacts:} OpenEarthMap models could enable automated mapping of any location on Earth, which can support decision making in disaster response, environmental conservation, and urban planning. However, such models will make it easy for anyone to access map information related to national security as well as privacy if sub-meter resolution images are available. Appropriate data analysis ethics and data policies are required to avoid security and privacy breaches.  

\vspace{-1mm}

\section*{Acknowledgement}
This work was supported by JST FOREST Grant Number JPMJFR206S, Japan.

\pagebreak

{\small
\bibliographystyle{ieeefullname}
\bibliography{wacv2023}
}
\clearpage
\pagebreak

\begin{strip}
\centering
\section*{Supplementary Material}
\end{strip}

	\supplementarysection
	In this supplementary, we provide a detailed description of the land cover classes of the OpenEarthMap dataset. We also present a brief overview of the baseline methods that were experimented on the OpenEarthMap dataset for the semantic segmentation and unsupervised domain adaptation tasks. The training settings and more experimental results on  the semantic segmentation and the unsupervised domain adaptation tasks are presented. Furthermore, we provide more land cover mapping results that were created from out-of-sample images (i.e., images not included in OpenEarthMap) to further demonstrate the generalization of the OpenEarthMap feature space. The attribution of all source data is summarized at the end.
	
	\section{Land Cover Classes}\label{sup-sec:1}
	Using the Anderson classification~\cite{anderson1976land} as a starting point, we subdivided the \textit{urban} class into three classes: \textit{building}, \textit{road}, and \textit{developed space}, which are visually interpretable in high-resolution images at a sub-meter level of ground sampling distance (GSD). The definitions of 8 classes in OpenEarthMap are summarized as follows.

\begin{figure}[t]
		\begin{center}
			\includegraphics[width=\linewidth]{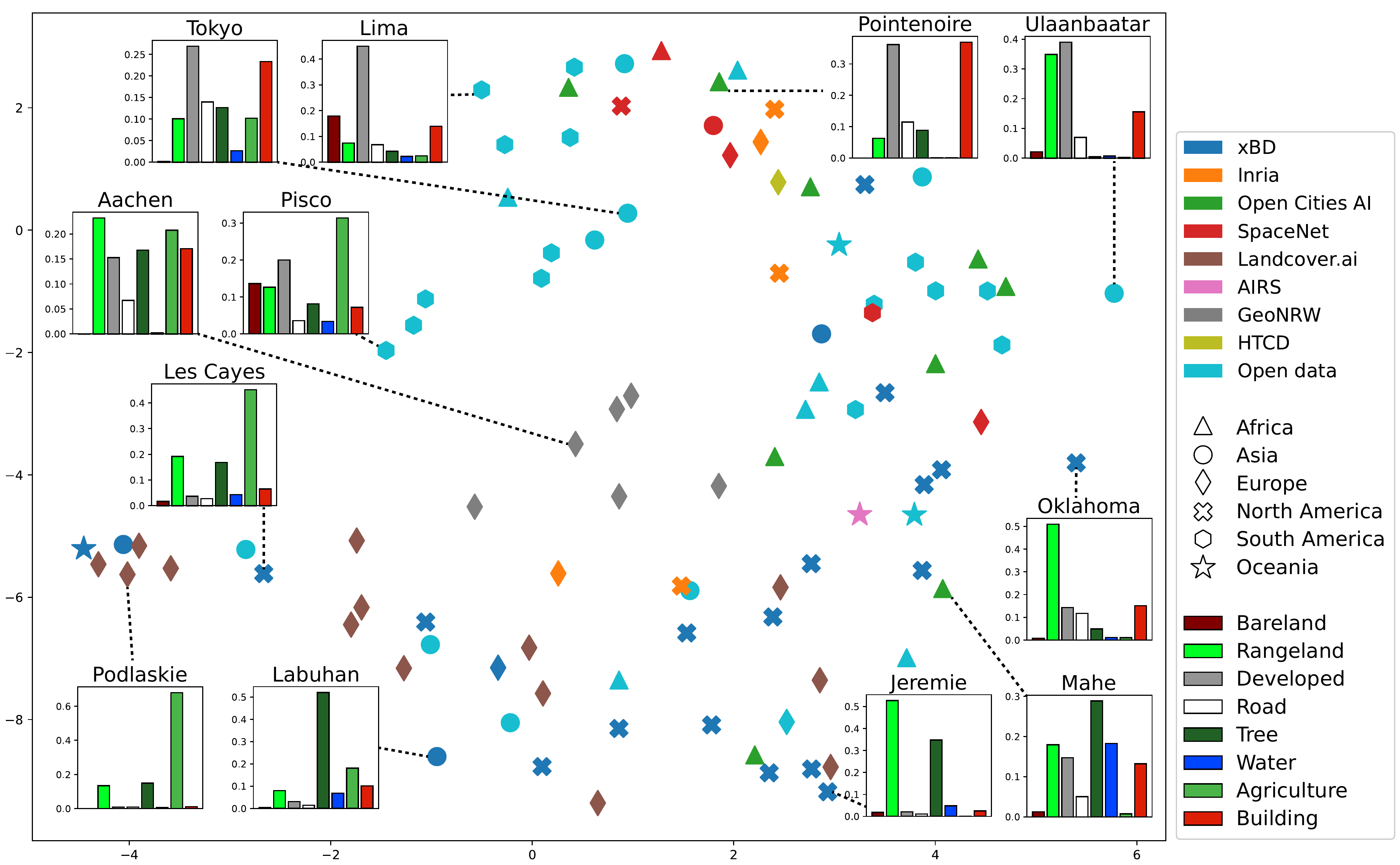}
		\end{center}
		\vspace{-3mm}
		\caption{t-SNE 2D visualization of the 97 regions based on the class proportions in the OpenEarthMap dataset.}
		\label{fig:tSNE_class}
	\end{figure}

	\begin{itemize}
		\item \textbf{Bareland} includes natural areas covered by sand or rocks without vegetation, and other accumulations of earthen materials.
		\item \textbf{Rangeland} includes areas dominated by herbaceous vegetation or bushes that are not cultivated or grazed, as well as grass and shrubs in gardens, parks, and golf courses.
		\item \textbf{Developed space} includes areas such as sidewalks, pavements, footpaths, parking lots, and construction sites as well as artificial grass areas search as tennis courts, baseball and football fields, etc. A lane in-between parking lots is considered as a road. The materials include asphalt, concrete, stones, bricks, and tiles. Compacted soil is also labeled as developed space.
		\item \textbf{Road} includes lanes, streets, railways, airport runways, and highway/motorway  for vehicles (e.g., trucks, cars, motorbikes, trains, and airplanes) excluding bicycles. The materials of roads include asphalt, concrete, and soil.
		\item \textbf{Tree} includes individual trees and a group of trees  that are identified from their shapes (shadow) and height.
		\item \textbf{Water} includes water bodies (e.g., rivers, streams, lakes, sea, ponds, dams) and swimming pools.
		\item \textbf{Agriculture land} includes areas used for producing crops (e.g., rice, wheat corn, soybeans, vegetables, tobacco, and cotton), perennial woody crops (e.g., orchards and vineyards), and non-native vegetation for grazing.
		\item \textbf{Building} includes residential, commercial, and industrial buildings.
	\end{itemize}
	Figure~\ref{fig:tSNE_class} shows a t-SNE 2D plot of the 97 regions using class proportions in the OpenEarthMap dataset. As can be seen in the bar graphs of 12 representative regions, the class proportions in the different regions are diverse.
	
	\begin{figure*}[!t]
		\
		\begin{center}
			\includegraphics[width=\linewidth]{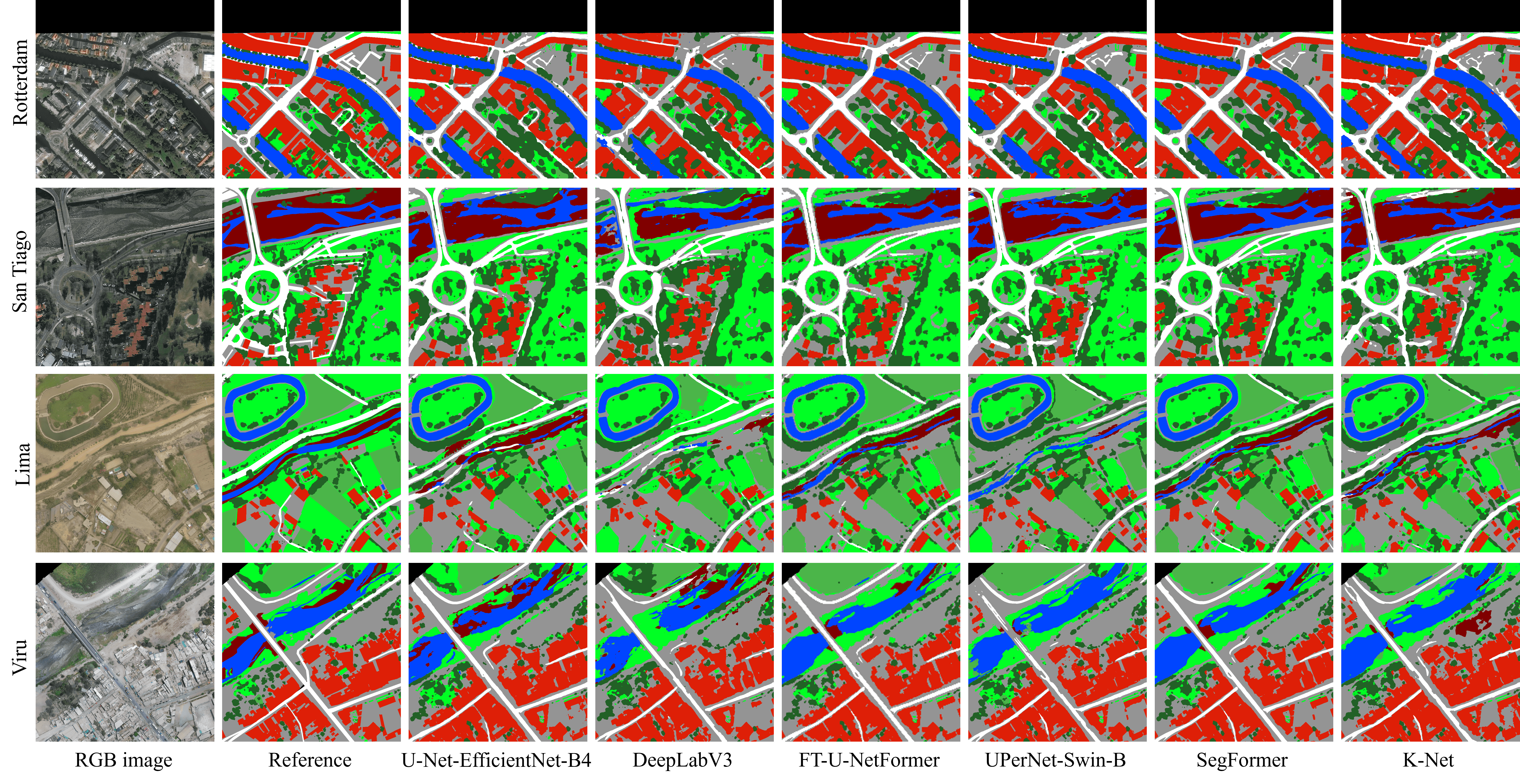}
		\end{center}
		\vspace{-3mm}
		\caption{Visual comparison of land cover mapping results of some of the baseline models.}
		\label{fig:segmentation_results_supp_compressed}
	\end{figure*}
	
	\section{Land Cover Semantic Segmentation}\label{sup-sec:2}
	\subsection{Brief Overview  of the Baselines}\label{sup-sec:2.1}
	The land cover semantic segmentation baseline networks that were experimented on the OpenEarthMap dataset are CNN-based (U-Net~\cite{ronneberger2015u}, DeepLabV3~\cite{deeplabv3plus2018}, HRNet~\cite{SunXLW19},  K-Net~\cite{zhang2021knet}, and ConvNeXt~\cite{liu2022convnet}) and Transformer-based (U-NetFormer~\cite{WANG2022196}, FT-U-NetFormer~\cite{WANG2022196}, SETR~\cite{SETR}, SegFormer~\cite{xie2021segformer}, UPerNet~\cite{xiao2018unified} with the backbones of ViT~\cite{dosoViTskiy2020}, Twins~\cite{chu2021twins}, and Swin Transformer~\cite{liu2021Swin}) architectures. U-Net uses an encoder-decoder structure to extract objects and image context at different scales. 
	U-Net models with VGG-11, ResNet-34, and EfficientNet-B4 as backbones were adopted. DeepLabV3~\cite{deeplabv3plus2018} uses a dilation hyperparameter of convolutional layers to develop atrous spatial pyramid pooling for robust object segmentation through many scales. HRNet builds high-resolution representations by continually executing multi-scale fusions across parallel convolutions. We employed an HRNet with W48 as a backbone. K-Net~\cite{zhang2021knet} separates instances and semantic categories uniformly with a number of learnable kernels. The kernels conduct convolution on the image features to provide segmentation predictions. ConvNeXt~\cite{liu2022convnet} is a pure ConvNet model constructed entirely from standard ConvNet modules. UPerNet with a backbone of ConvNeXt-B was used. The U-NetFormer~\cite{WANG2022196} selects the advanced ResNext101 as the encoder
	and develops an efficient global–local attention mechanism to model both global and local information in the
	decoder. FT-U-NetFormer~\cite{WANG2022196} replaces the CNN encoder with the
	Swin Transformer (Swin-B).
	SETR~\cite{SETR} interprets an input image as a sequence of patches represented by a learned patch embedding, then, modifies the sequence using a global self-attention module for discriminative feature representation learning. A SETR PUP with a backbone of ViT-L was used. SegFormer~\cite{xie2021segformer} unifies transformers with lightweight multilayer perceptron decoders without considering positional encoding. A MiT-B5 encoder was used as a backbone for SegFormer. Swin Transformer~\cite{liu2021Swin} creates hierarchical feature maps by merging image patches into deeper layers. Twins~\cite{chu2021twins} uses a spatially separable attention mechanism comprising of locally-grouped self-attention and global sub-sampled attention.

	\begin{table}[!t]
		\caption{Performance of ImageNet pre-trained model with different optimizers.}
		\vspace{-5mm}
		\begin{center}
			\begin{tabular}{cccc}
				\hline \hline
				Model	&	Pre-trained on	&	Optimizer	&	mIoU	\\
				\hline
				\multirow{4}{*}{UPerNet-Swin-B}	&	ImageNet	&	SGD	&	62.15	\\
				&	ImageNet	&	AadmW	&	66.09	\\
				&	None	&	SGD	&	62.32	\\
				&	None	&	AadmW	&	\textbf{66.13}	\\
				\hline \hline
			\end{tabular}
		\end{center}
		\label{tab:pretraining_results}
	\end{table}
	
	\begin{table}[!t]
		\vspace{-3mm}
		\caption{The results of using different input patch sizes and different loss functions with UPerNet-Swin-B.}
		\vspace{-2mm}
		\begin{subtable}[c]{0.2\textwidth}
			\subcaption{Input patch size}
			\centering
			\begin{tabular}{cc}
				\hline\hline
				Patch size	&	mIoU	\\
				\hline
				512$\times$512	&	66.09	\\
				620$\times$620	&	66.14	\\
				768$\times$768	&	66.17	\\
				1024$\times$1024	&	\textbf{67.02}	\\
				\hline \hline
			\end{tabular}
		\end{subtable}
		\begin{subtable}[c]{0.2\textwidth}
			\subcaption{Loss functions}
			\centering
			\begin{tabular}{cc}
				\hline \hline
				Function	&	mIoU	\\
				\hline
				CE	&	66.09	\\
				CE+Lovasz	&	66.21	\\
				CE+Focal	&	65.89	\\
				CE+Lovasz+Focal	&	\textbf{66.38}	\\
				\hline\hline
			\end{tabular}
		\end{subtable}
		\label{tab:patch_pretraining_results}
	\end{table}

	\begin{table*}[!t]
		\captionsetup{width=.98\linewidth}
		\caption{Compact segmentation models discovered on OpenEarthMap training set with SparseMask and FasterSeg. The class IoUs and the mIoU are calculated on the test set of OpenEarthMap with TTA applied.}
		\vspace{-2.5mm}
		\begin{center}
			\scalebox{0.8}{
				\begin{tabular}{c c c c c c c c c c c c c c}
					\hline \hline
					\multirow{2}{*}{Method} & \multirow{2}{*}{Trial} & \multicolumn{8}{c}{IoU (\%)}  & {mIoU} & {Params} & {FLOPs} & {FPS}
					\\ \cline{3-10} 
					& & Bareland	&	Rangeland	&	Developed	&	Road	&	Tree	&	Water	&	Agriculture	&	Building	& (\%) & (M) & (G) & (ms) \\	\hline
					\multirow{2}{*}{SparseMask} & 1st & \textbf{47.00} &  \textbf{53.42} &  46.40 &  46.94 &  66.98 &  \textbf{79.02} &  72.64 &  69.25 &  \textbf{60.21} & 2.96 & \textbf{10.28} & 25.9 \\
					& 2nd & 45.96 & 53.01 &  \textbf{46.71} &  47.26 &  67.12 &  78.81 &  \textbf{71.88} &  69.27 &  60.00 & 3.10 & 10.39 & 26.4 \\
					\hline
					\multirow{2}{*}{FasterSeg} & 1st & 34.04 &  51.40 &  44.97 &  55.82 &  66.58 &  74.50 &  70.33 &  69.14 &  58.35 & \textbf{2.23} & 14.58 & 74.8 \\
					& 2nd & 35.78 &  52.03 &  46.32 &  \textbf{56.97} &  \textbf{67.20} &  75.76 &  70.70 &  \textbf{70.55} &  59.41 & 3.47 & 15.37 & \textbf{89.5} \\
					\hline \hline
			\end{tabular}}
		\end{center}
		\label{tab:nas_results1}
	\end{table*}
	
	\begin{table*}[!htt]
		\captionsetup{width=.98\linewidth}
		\caption{Compact segmentation models discovered on OpenEarthMap training set with SparseMask and FasterSeg. The class IoUs and the mIoU are calculated on the test set of OpenEarthMap without TTA applied.}
		\vspace{-2.5mm}
		\begin{center}
			\scalebox{0.8}{
				\begin{tabular}{c c c c c c c c c c c c c c}
					\hline \hline
					\multirow{2}{*}{Method} & \multirow{2}{*}{Trial} & \multicolumn{8}{c}{IoU (\%)}  & {mIoU} & {Params} & {FLOPs} & {FPS}
					\\ \cline{3-10} 
					& & Bareland	&	Rangeland	&	Developed	&	Road	&	Tree	&	Water	&	Agriculture	&	Building	& (\%) & (M) & (G) & (ms) \\	\hline
					\multirow{2}{*}{SparseMask} & 1st & \textbf{46.15} &  \textbf{51.88} &  44.01 &  43.64 &  65.20 &  \textbf{77.41} &  \textbf{71.47} &  66.10 &  58.23 & 2.96 & \textbf{10.28} & 51.2 \\
					& 2nd & 44.78 &  51.56 &  44.34 &  43.95 &  65.44 &  77.34 &  70.98 &  66.09 &  58.06 & 3.10 & 10.39 & 57.2 \\
					\hline
					\multirow{2}{*}{FasterSeg} & 1st & 33.52 &  50.60 &  43.93 &  54.74 &  65.98 &  73.55 &  69.73 &  68.36 &  57.55 &\textbf{ 2.23} & 14.58 & 143.2 \\
					& 2nd & 34.50 &  51.27 &  \textbf{45.27} &  \textbf{55.94} &  \textbf{66.61} &  74.71 &  70.05 &  \textbf{69.73} &  \textbf{58.51} & 3.47 & 15.37 & \textbf{171.3} \\
					\hline \hline
			\end{tabular}}
		\end{center}
		\label{tab:nas_results2}
		\vspace{0mm}
	\end{table*}
	
	\subsection{Experimental Details}
	All the baselines we used for the experiments are PyTorch-based. The U-Net-based architectures are adopted from Yakubovskiy~\cite{Yakubovskiy:2019} and Wang \textit{et al.}~\cite{WANG2022196}, and the other architectures are from MMsegmentation~\cite{mmseg2020}. The networks were trained on a single NVIDIA GPU DGX-1/DGX-2 with 16/32GB of RAM. The number of epochs was set to 200, and a batch size of 8 with an image input size of 512$\times$512 randomly cropped was employed. The cross-entropy (CE) loss was used in training all the networks. For the U-Net-based architectures, we used AdamW optimizer~\cite{loshchilov2018decoupled} with a learning rate of $1\times10^{-4}$ and weight decay of $1\times10^{-6}$. For the MMsegmentation-based architectures, we used the default settings of each method. We adopted stochastic gradient descent (SGD) optimizer with a learning rate of $1\times10^{-3}$, weight decay of $5\times10^{-4}$, and momentum of 0.9 for the DeeplabV3 and HRNet networks. The rest of the networks used AdamW optimizer with a learning rate set as $6\times10^{-5}$, weight decay as 0.01, and betas parameters as 0.9 and 0.999. A polynomial learning rate decay with a factor of 1.0 and an initial linear warm-up of 1500 iterations was used. The backbones in all the networks were pre-trained on the ImageNet dataset. No data augmentation was applied during training for all networks. Following previous works, we used mIoU to assess the performance of all models. All results are based on test-time augmentation (TTA) with flipping operations. 
	
	\subsection{Results}\label{sup-sec:2.2}
	\noindent\textbf{Visualization:}
	More visual examples of segmentation results obtained from some selected baseline methods are presented in Figure~\ref{fig:segmentation_results_supp_compressed}. 
	In the first row (Rotterdam), DeepLabV3 failed to identify the entire stretch of \textit{road}. In the second row (San Tiago), the entire \textit{bareland} and the small \textit{developed space} on top of the \textit{bareland} were identified by FT-U-NetFormer and SegFormer. U-Net-EfficientNet-B4,  FT-U-NetFormer and SegFormer were able to identify the \textit{bareland} that stretches from along the river (third row), but DeepLabV3, UPerNet-Swin-B, and K-Net failed to identify it. Compared to DeepLabV3 and K-Net, U-Net-EfficientNet-B4, DeepLabV3, and FT-U-NetFormer did recognize most parts of the \textit{water} body in Viru (fourth row). UPerNet-Swin-B and DeeplabV3 classified the \textit{water} body in Viru as \textit{agricultural land}, \textit{rangeland}, and \textit{developed space}.
	
	\noindent\textbf{Ablation study:} 
	We conducted an ablation study to examine the effects of ImageNet pre-training, optimizers, image size, and loss functions on the OpenEarthMap dataset using a UPerNet-Swin-B network. We employed two different optimizers (AadmW and SGD), three  loss functions (CE, Lovasz, and Focal), and four different patch sizes. As shown in Table~\ref{tab:pretraining_results}, the model that was not pre-trained on ImageNet performed better than the one pre-trained on ImageNet. In both cases, AdamW optimizer attains better results. Table~\ref{tab:patch_pretraining_results} shows the results of using different patch sizes and different loss functions for training UPerNet-Swin-B. A larger patch size tends to achieve better performance. The combination of all three loss functions (CE, Lovasz, and Focal) can improve the performance of UPerNet-Swin-B. \par

	\subsection{Neural Architecture Search Methods}\label{sup-sec:2.3}
	The two automated neural architecture search methods, SparseMask~\cite{Wu2019SparseMask} and FasterSeg~\cite{Chen2020FasterSeg}, which we adopted were particularly proposed for compact architecture search for semantic segmentation tasks in computer vision. Both methods employed a gradient-based search strategy similar to the one used in DARTS~\cite{liu2018darts}. Whereas SparseMask used a pruning technique to compress the searched architectures, teacher-student co-searching (knowledge distillation technique) was used in FasterSeg. Following the architecture search protocol in both methods, we conducted four experiments, two with each method, by searching for lightweight architectures on the OpenEarthMap dataset. All the experiments we performed on a single NVIDIA Tesla P100 with 16GB memory. It took about 0.8 GPU days and 2 GPU days to perform the architecture search with SparseMask and FasterSeg, respectively. 
	The architectures discovered with both methods were trained from scratch using the same training protocol adopted in \cite{Wu2019SparseMask} and \cite{Chen2020FasterSeg} with the exception of setting the number of epochs to 450. We measured the class-specific IoUs and mIoU using a single-scale input with TTA (see Table~~\ref{tab:nas_results1}) and without TTA (see Table~~\ref{tab:nas_results2}) of flipping operations applied to the test set of the OpenEarthMap dataset. The FPS and the FLOPs were calculated with a single image of 1024$\times$1024 pixels as an input. The inference speed was calculated using the settings in FasterSeg. Table~~\ref{tab:nas_results1} and Table~\ref{tab:nas_results2} show the detail class-specific IoUs along with the mIoUs that are presented in the paper.
		
	\begin{figure*}[!htt]
		\begin{center}
			\includegraphics[width=\linewidth]{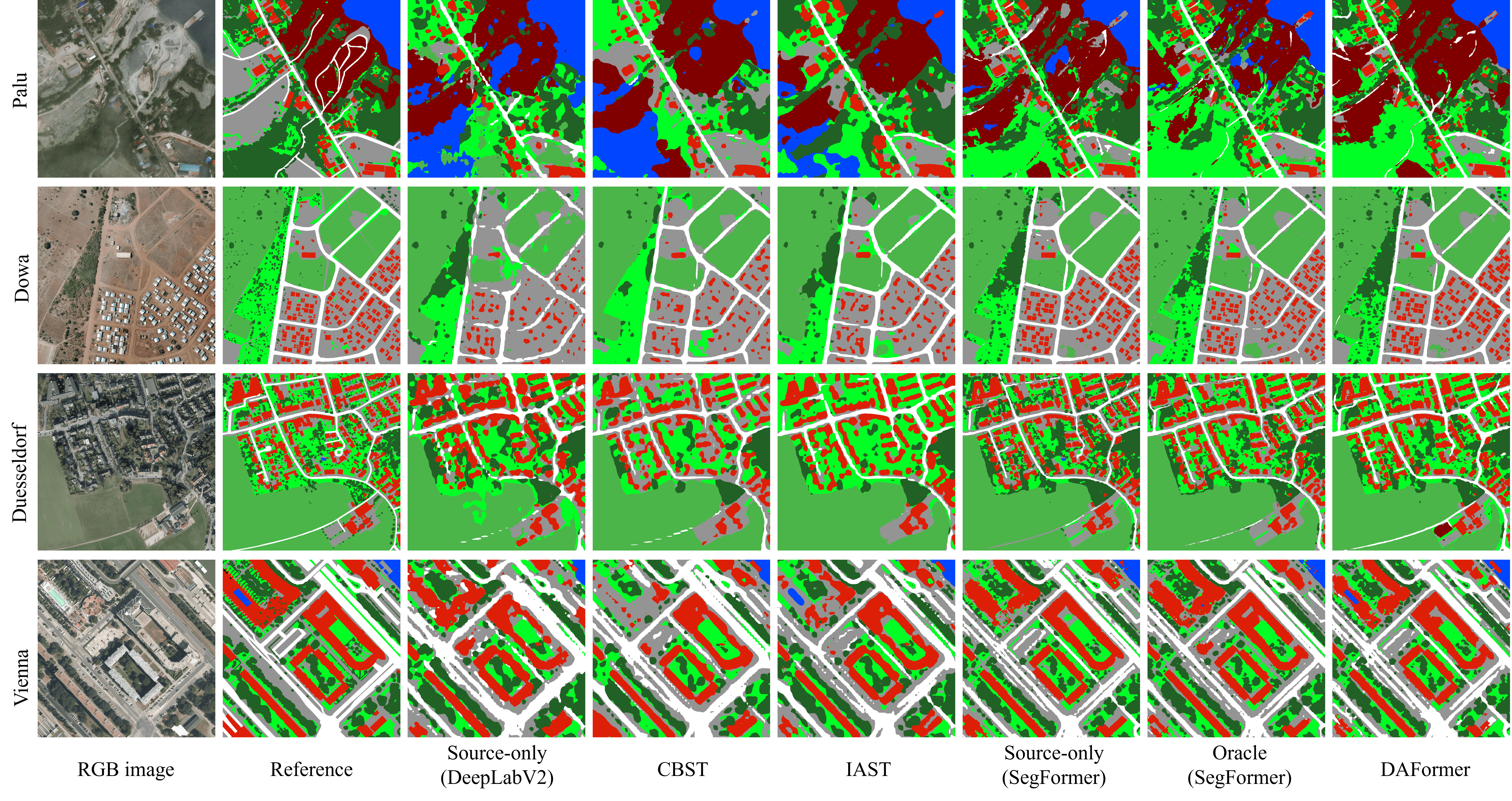}
		\end{center}
		\vspace{-5mm}
		\caption{Visual comparison of unsupervised domain adaption results of some of the baseline models.}
		\label{fig:uda_results_sup_compressed}
	\end{figure*}
	
	\begin{figure*}[!t]
		\centering
		\begin{minipage}[b]{\textwidth}
			\includegraphics[width=\linewidth]{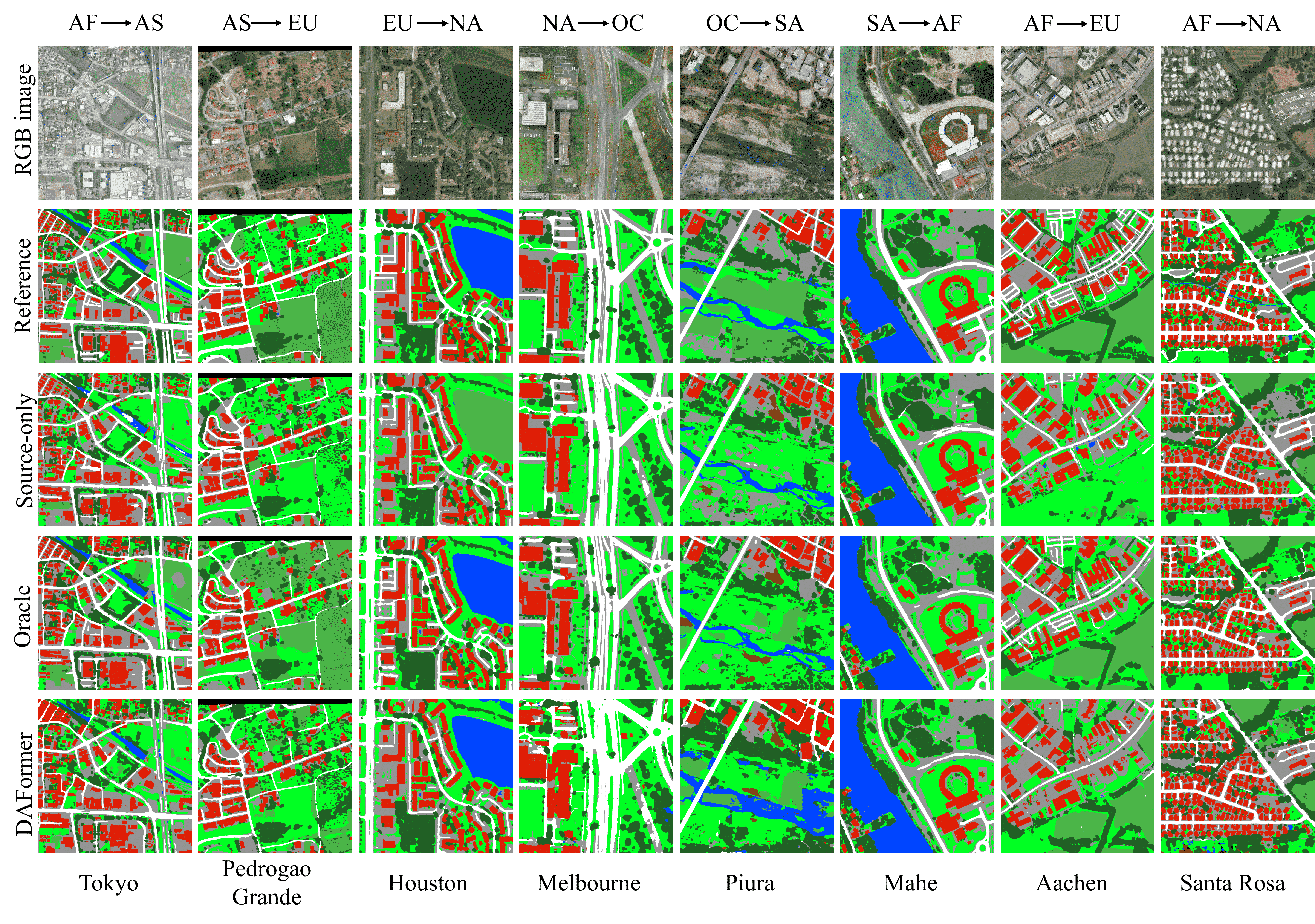}
			\vspace{-7mm}
			\caption{Visual comparison of continent-wise unsupervised domain adaption results of source-only SegFormer, Oracle and DAFormer. Asia: AS, Europe: EU, Africa: AF, North America: NA, South America: SA and Oceania: OC.}
			\label{fig:uda_results_contient_supp}
			\vspace{5mm}
		\end{minipage}
	\end{figure*}
	
	\begin{figure*}[!t]
		\centering
		\begin{minipage}[b]{.47\textwidth}
			\includegraphics[width=\linewidth]{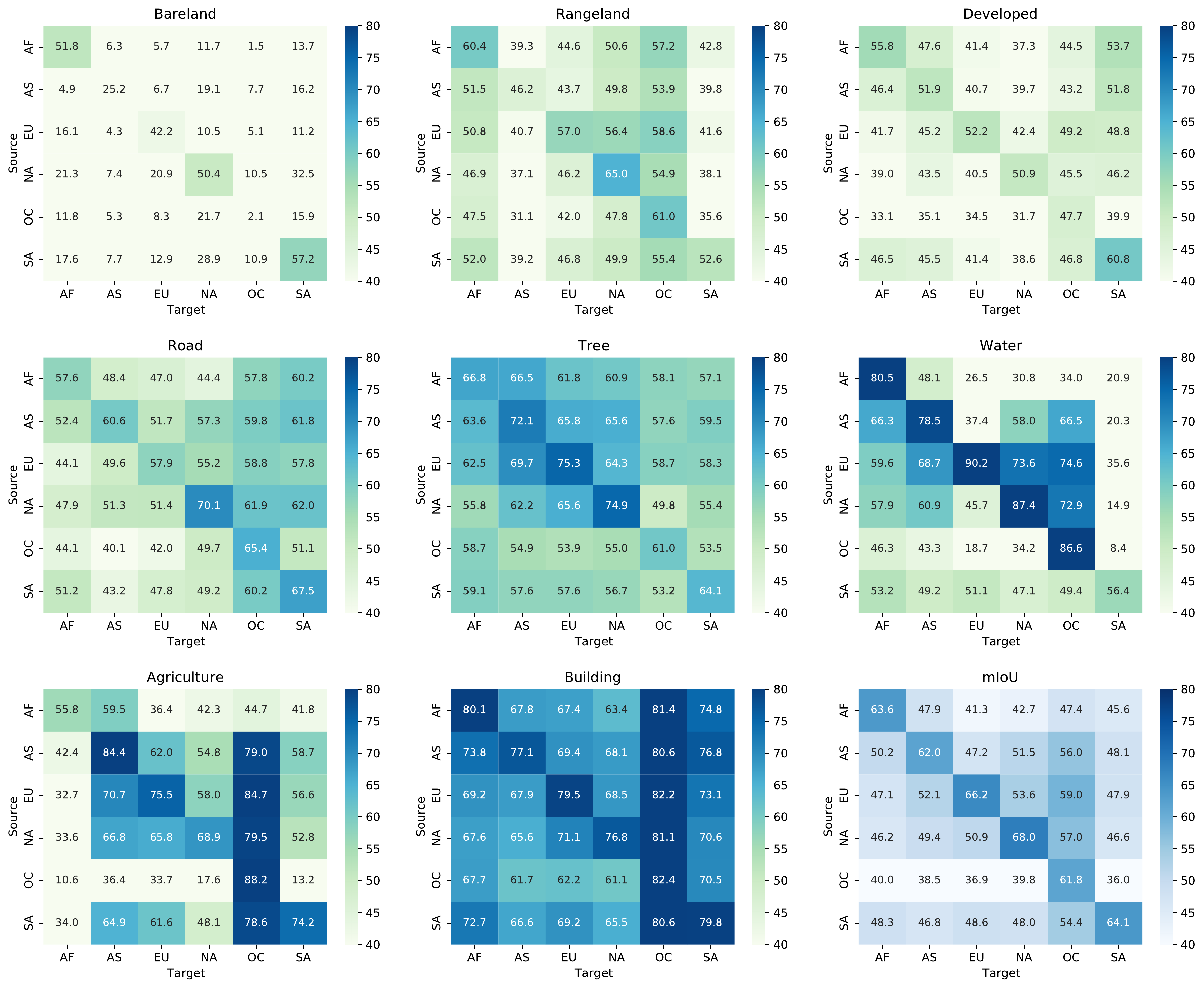}
			\caption{Continent-wise UDA results from source-only U-Net-EfficientNet-B4. Class-specific IoUs and mIoU are shown in the subfigures with oracle values in the diagonal.}
			\label{fig:sup_unet_continents}
		\end{minipage}\qquad
		\begin{minipage}[b]{.47\textwidth}
			\includegraphics[width=\linewidth]{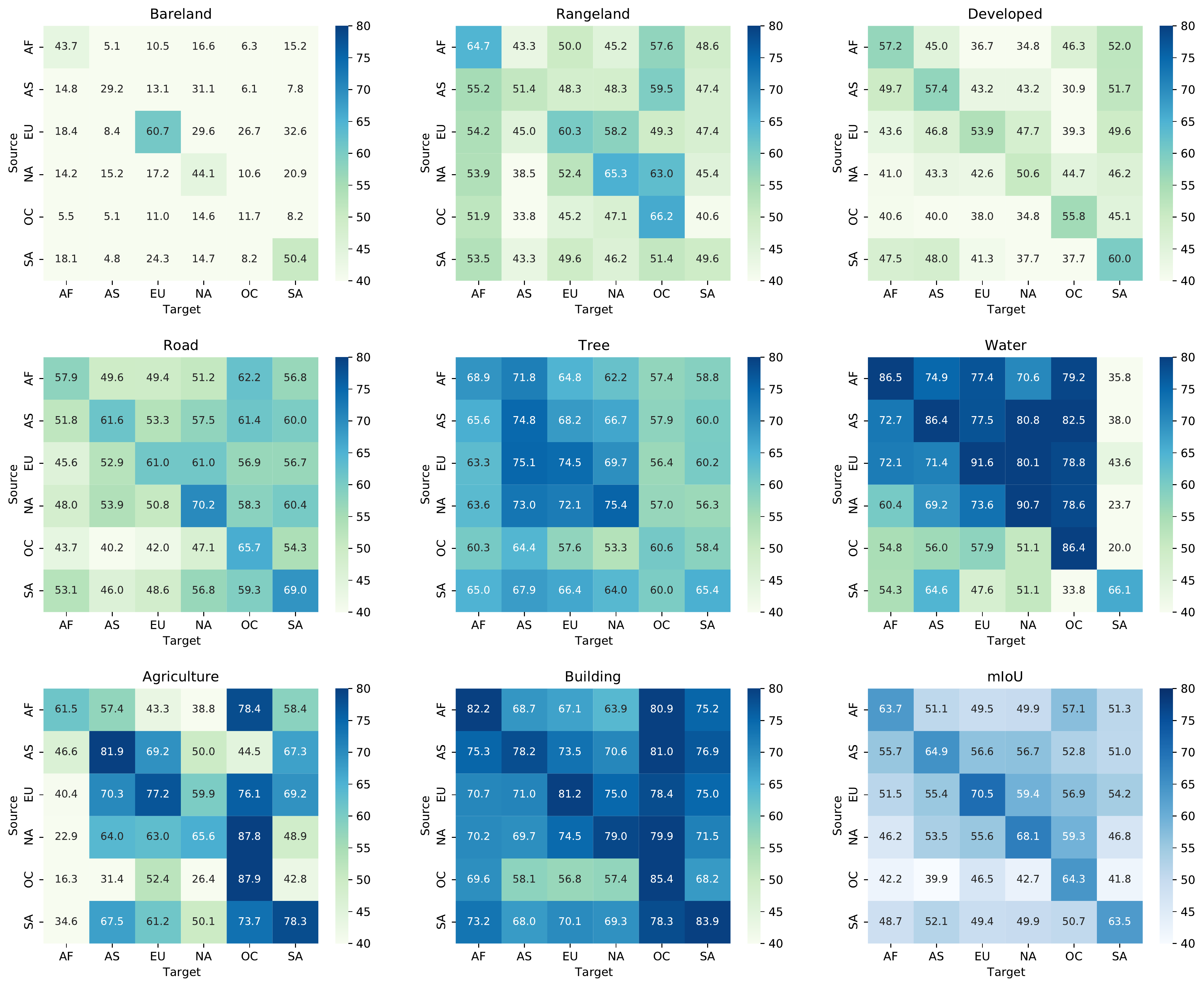}
			\caption{Continent-wise UDA results from source-only SegFormer. Class-specific IoUs and mIoU are shown in the subfigures with oracle values in the diagonal.}
			\label{fig:sup_segformer_continents}
		\end{minipage}
		\vspace{6mm}
	\vspace{5mm}
		\centering
		\begin{minipage}[b]{.47\textwidth}
			\includegraphics[width=\linewidth]{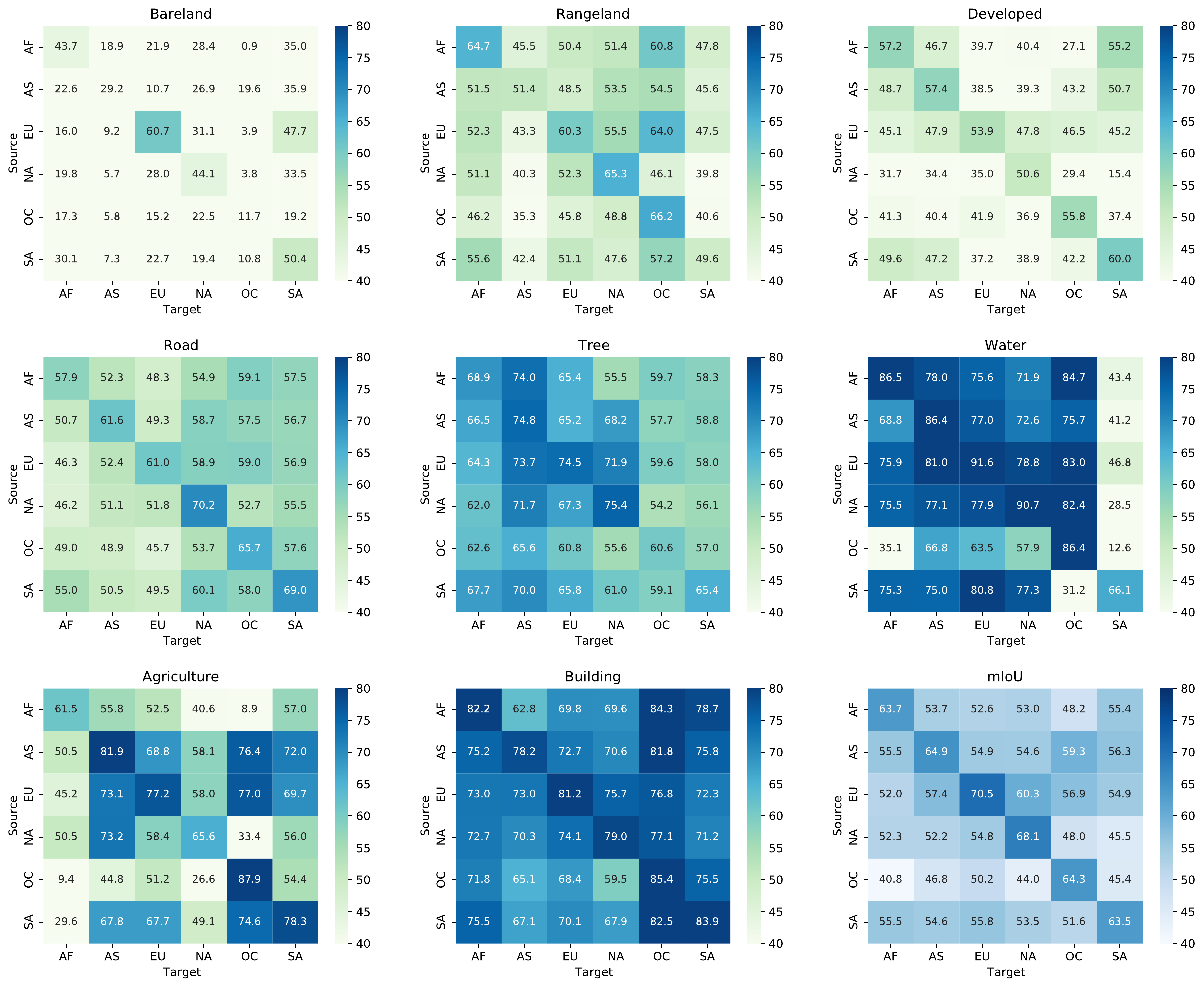}
			\caption{Performance of DAFormer in Continent-wise UDA. Class-specific IoUs and mIoU are shown in the subfigures with oracle values in the diagonal.}
			\label{fig:sup_daformer_continents}
		\end{minipage}\qquad
		\begin{minipage}[b]{.47\textwidth}
			\includegraphics[width=\linewidth]{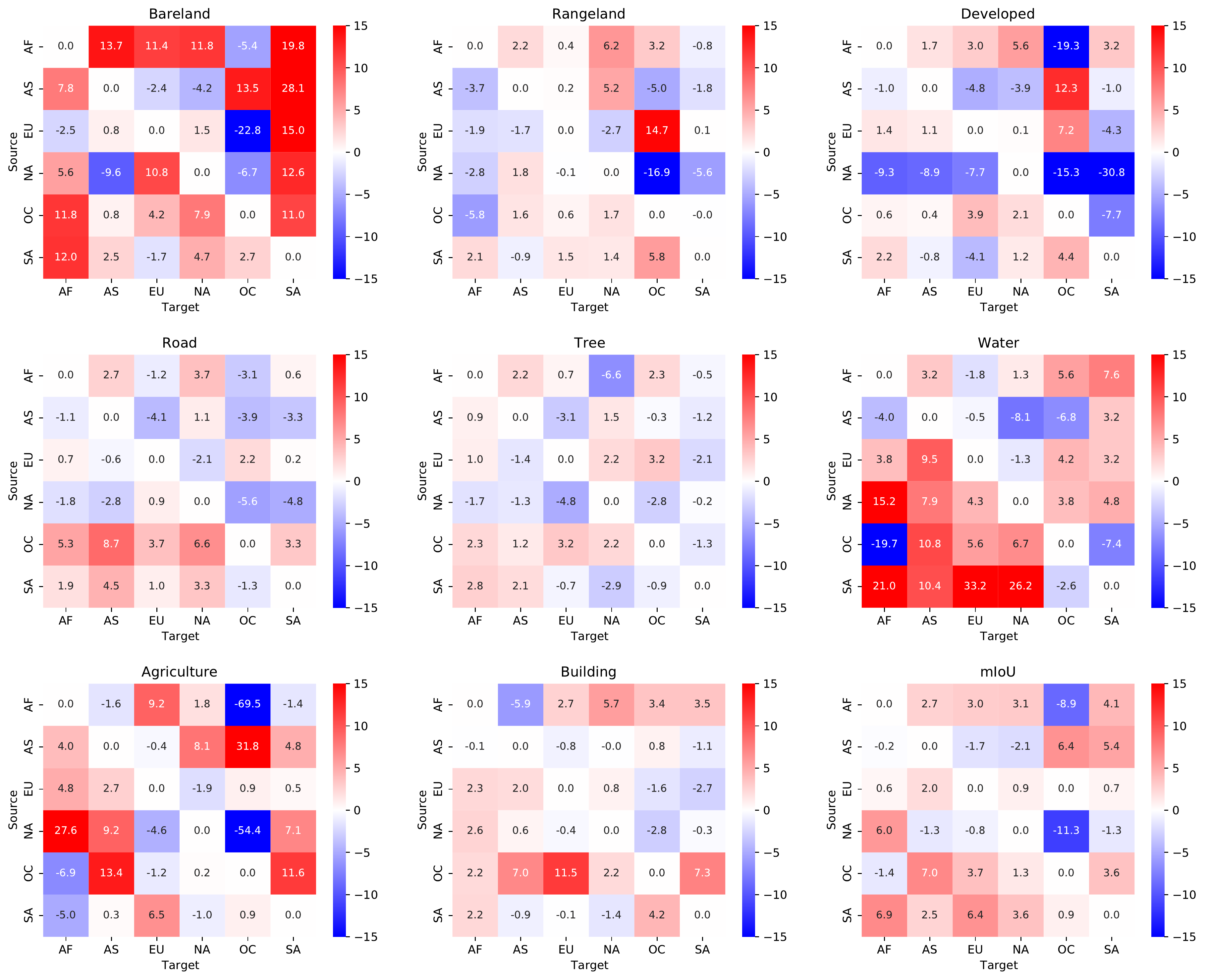}
			\caption{Performance change from SegFormer to DAFormer in Continent-wise UDA. Class-specific IoUs and mIoU differences are shown in the subfigures.}
			\label{fig:sup_difference_continents}
		\end{minipage}
	\end{figure*}

	\vspace{2mm}
	\section{Unsupevised Domain Adaptation}\label{sup-sec:3}
	\subsection{Brief Overview of the Baselines}\label{sup-sec:3.1}
	We adopted a metric-based method (MCD~\cite{TzengHZSD14}), adversarial training methods (AdaptSeg~\cite{Tsai_adaptseg_2018}, category-level adversarial network (CLAN)~\cite{luo2019Taking}, TransNorm~\cite{Wang19TransNorm}, and fine-grained adversarial learning framework for domain adaptive (FADA)~\cite{Haoran_2020_ECCV}), and self-training methods (pyramid curriculum DA (PyCDA)~\cite{Lian_2019_ICCV}, class-balanced self-training (CBST)~\cite{zou2018unsupervised}, instance adaptive self-training (IAST)~\cite{mei2020instance}, and DAFormer~\cite{hoyer2022daformer}) for the unsupervised domain adaptation task. DAFormer is a transformer-based model, and the others are based on DeepLabV2. The adversarial training methods seek to match the distributions of the source and target domains from input-feature-output or a patch level in a generative adversarial network. AdaptSeg~\cite{Tsai_adaptseg_2018} uses adversarial learning in the output space and a multi-level adversarial network to effectively perform output space domain adaptation at different feature levels. CLAN~\cite{luo2019Taking} aligns the classes with an adaptive adversarial loss. TransNorm~\cite{Wang19TransNorm} uses an end-to-end trainable layer to make networks more transferable across domains. FADA~\cite{Haoran_2020_ECCV} uses a fine-grained adversarial learning framework by aligning the class-level features. Self-learning methods generate the target domain's pseudo labels, retrain the model and repeat the procedure. PyCDA~\cite{Lian_2019_ICCV} observes the target properties and fuses multi-scale features. CBST~\cite{zou2018unsupervised} and IAST~\cite{mei2020instance} aim at selecting balanced samples to improve the quality of pseudo labels. DAFormer~\cite{hoyer2022daformer} constructs a transformer encoder and a multilevel context-aware feature fusion decoder by adopting three effective training strategies: rare class sampling, ImageNet Feature Distance, and a learning rate warmup. \par
	
	\begin{table}[!t]
		\small
		\caption{Comparing source-only and oracle training for different networks on the 24 regions test set of OpenEarthMap.}
		\vspace{-5mm}
		\begin{center}
			\begin{tabular}{ccc}
				\hline \hline
				\multirow{2}{*}{Model} & \multicolumn{2}{c}{mIoU (\%)} \\
				\cline{2-3} & Source-only &	Oracle	\\
				\hline
				U-Net-EfficientNet-B4 & \textbf{63.17} & 64.09\\
				DeepLabV2 & 50.01 & 54.65 \\
				DeepLabV3 & 55.27 & 59.83\\
				HRNet & 56.25& 60.02 \\
				SegFormer & 58.25&\textbf{64.76} \\
				UPerNet-Swin-B & 52.35	& 61.82	\\	
				K-Net & 57.21 & 64.12 \\
				\hline \hline
			\end{tabular}
		\end{center}
		\vspace{-6mm}
		\label{tab:UDA_networks}
	\end{table}
	
	\subsection{Experimental Details}
	For the DeepLabV2-based methods, we adopted the architectures in  Wang \textit{et al.}~\cite{wang2021loveda} and kept the default setting. We used the following training settings (batch size of 8 with image input size of 512×512 randomly cropped) in the semantic segmentation task for all the DeepLabV2-based methods and the DAFormer. 
	DeepLabV2 with ResNet50 was used as an extractor, and a discriminator was constructed using fully convolutional layers. 
	The classification and the discriminator learning rates for the adversarial training methods were set to $5 \times 10^{-3}$ and $10^{-4}$, respectively. Adam optimizer was used in the discriminator with momentum of $0.9$ and $0.99$. We adopted the default pseudo-generation hyper-parameters in CBST and IAST, and set the classification learning rate to $10^{-2}$. All the networks were trained for 40K steps in two stages. In the first stage, the models were trained only on the source images for 8K steps for initialization. In the second stage, the pseudo-labels were then updated every 2K steps for the remaining training process. For the DAFormer, an MiT-B5 encoder was adopted with AdamW. Other hyper-parameters remained the same as in the original literature~\cite{hoyer2022daformer}.

	\subsection{Results}\label{sup-sec:3.2}
	\noindent\textbf{Visualization:} More visual examples of the UDA results are presented in Figure~\ref{fig:uda_results_sup_compressed}. 
	In the first row (Palu), the source-only DeepLabV2 can barely identify the \textit{tree} (bottom-left) and \textit{developed space} (center-left). These areas were classified as \textit{bareland} and \textit{water}. CBST and IAST also did not perform well in these areas. Source-only SegFormer and DAFormer slightly improved their performance in these areas. In the second row (Dowa), source-only DeepLabV2 could not recognize the tiny \textit{road} and small \textit{buildings}. IAST and CBST performed better on the \textit{road}, but they cloud not recognize the small \textit{buildings}. DAFormer performs exceptionally well in those two areas. In the third row (Dusseldorf), DAFormer could identify the long and tiny \textit{road} (bottom), and the other methods only recognized some parts of the \textit{road}.
	The IAST did identify the small \textit{water} area (top-left) in Vienna (fourth row), and the other methods classified it as \textit{building}.

	\noindent\textbf{Comparison of network architectures:}
	To further evaluate the suitability of the networks for the UDA task, we performed several experiments with source-only and oracle training for different models and provided their mIoU results in Table.~\ref{tab:UDA_networks}. The classical DeepLabV2 yielded the worst results in both source-only and oracle. UPerNet-Swin-B was slightly better than DeepLabV2 in the source-only setting. SegFormer obtained the best oracle performance, and U-Net-EfficientNet-B4 outperformed SegFormer on the source-only setting. 
	Generally, U-Net-EfficientNet-B4, SegFormer, and K-Net shared the top positions in both source-only and oracle settings. These networks are recommended to be further investigated for the development of UDA methods on the OpenEarthMap dataset. \par

	\noindent\textbf{Continent-wise UDA:} 
	Visual comparison of continent-wise UDA results of source-only SegFormer, Oracle and DAFormer are shown in Figure~\ref{fig:uda_results_contient_supp}. Eight combinations of source and target domains are provided. Here, we denote Africa, Asia, Europe, North America, South America, and Oceania as AF, AS, EU, NA, SA, and OC, respectively. For AF$\rightarrow$AS, AS$\rightarrow$EU, and AF$\rightarrow$EU, DAFormer significantly achieved better results on \textit{agriculture land} when compared to source-only SegFormer. DAFormer also identified the \textit{water} in EU$\rightarrow$NA, and the small \textit{buildings} and the tiny \textit{roads} in SA$\rightarrow$AF and AF$\rightarrow$NA. The class-specific IoUs and mIoUs obtained from source-only U-Net-EfficientNet-B4, source-only SegFormer, and DAFormer are presented in  Figure~\ref{fig:sup_unet_continents}, Figure~\ref{fig:sup_segformer_continents}, and Figure~\ref{fig:sup_daformer_continents}, respectively. For most classes, when OC is considered as the source domain, the transferred result is worst due to the limited number of images in OC. However, when OC is treated as the target domain the performance is better than other settings. With the exception of OC, \textit{building} is the easiest transferred class, which U-Net-EfficientNet-B4 achieved IoUs range of 63.4 to 76.8 and SegFormer achieved a range of 64.3 to 76.3. The most challenging transferred class is \textit{bareland}. The U-Net-EfficientNet-B4 and the SegFormer attained IoUs range of 4.3 to 32.5 and 4.1 to 32.7, respectively. In the order of easiest to challenging class is \textit{building}, \textit{tree}, \textit{road}, \textit{rangeland}, \textit{developed space},\textit{water}, \textit{agriculture land}, and \textit{bareland}. The performance change from SegFormer to DAFormer is shown in Figure~\ref{fig:sup_difference_continents}. Red indicates an improvement whereas blue depicts a decrease in results. In Figure~\ref{fig:sup_difference_continents}, one can clearly see that DAFormer improved the results in the challenging classes (e.g., \textit{bareland} and \textit{water}).

	\section{Mapping for Out-of-Sample Images}\label{sup-sec:4}
\begin{figure*}[!t]
		\centering
		\begin{minipage}[b]{1\textwidth}
			\includegraphics[width=\linewidth]{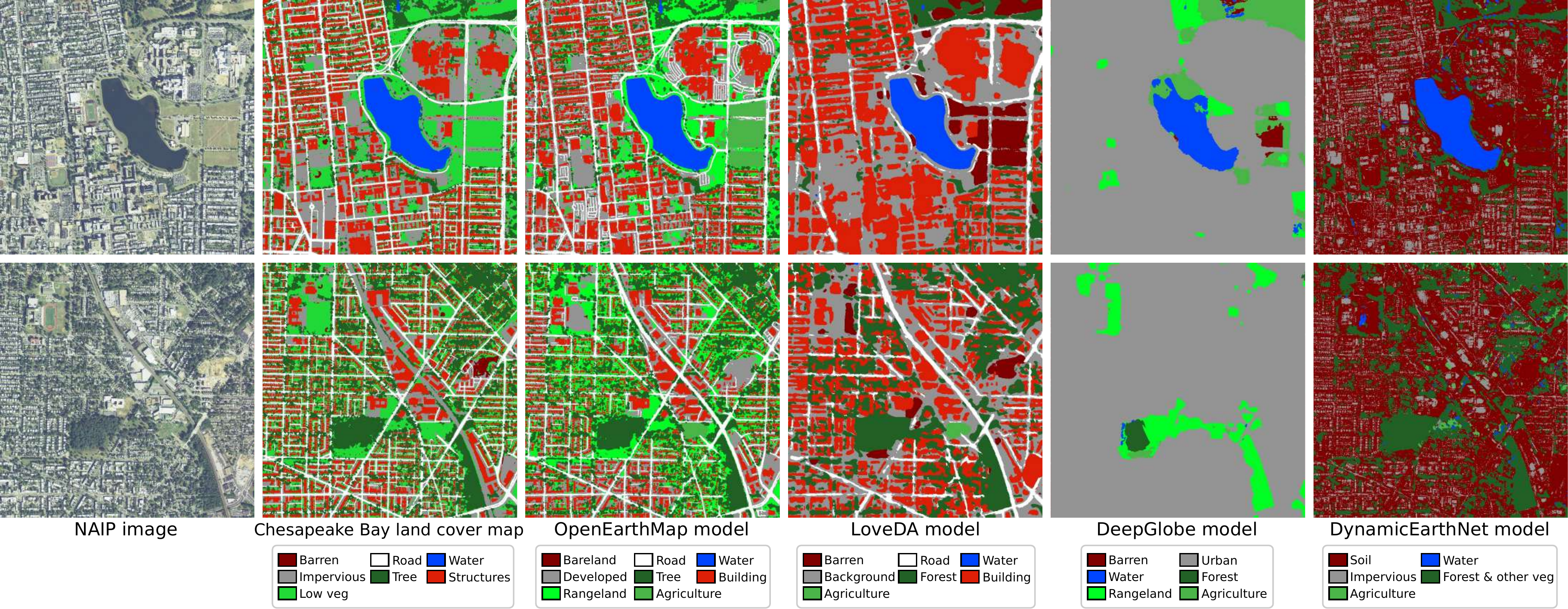}
			\vspace{-4mm}
			\caption{Visual comparison of Chesapeake Bay land cover map with land cover maps generated by U-Net models trained on OpenEarthMap, LoveDA, DeepGlobe, and DynamicEarthNet. The NAIP images are the source data.}
			\label{fig:chesapeake}
			\vspace{6mm}
		\end{minipage}\hfil
		\begin{minipage}[b]{1\textwidth}
			\centering
			\includegraphics[width=0.83\linewidth]{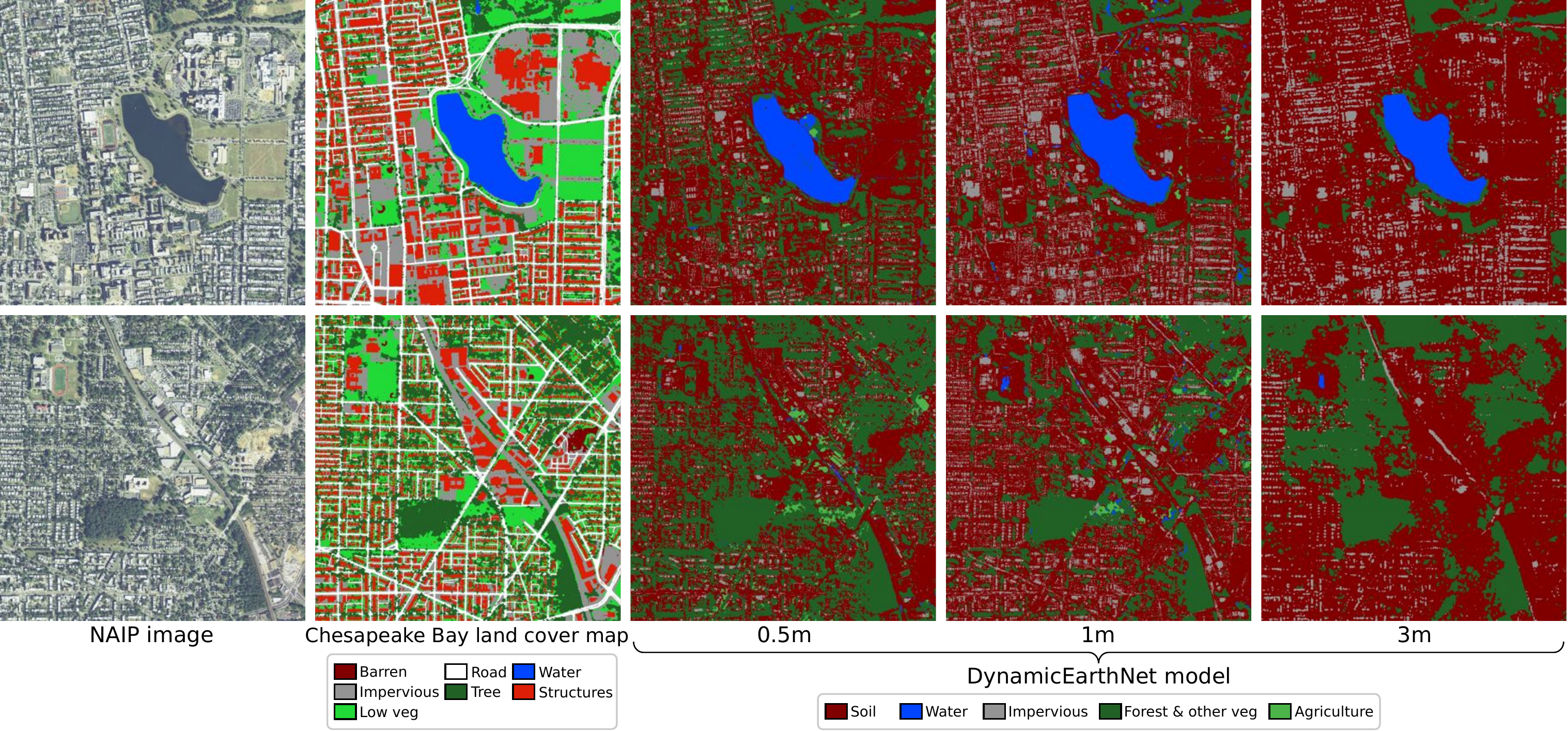}
			\vspace{-2mm}
			\caption{Visual comparison of land cover maps generated by U-Net trained on DynamicEarthNet using NAIP images at different GSDs (i.e., 0.5cm, 1m, and 3m) for inference.}
			\label{fig:chesapeake2}
			\vspace{6mm}
		\end{minipage}\hfil
		\begin{minipage}[b]{1\textwidth}
			\includegraphics[width=\linewidth]{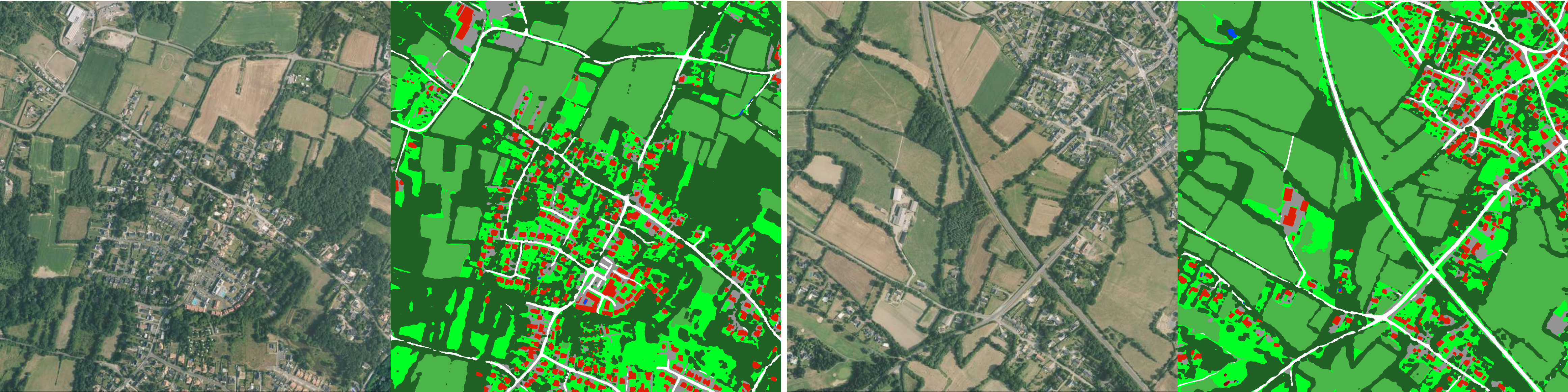}
			\vspace{-3mm}
			\caption{Out-of-sample mapping examples of MiniFrance from France.}
			\label{fig:minifrance}
		\end{minipage}
	\end{figure*}

	Figure~\ref{fig:chesapeake} shows visual comparisons of the Chesapeake Bay land cover map with those generated by U-Net-EfficientNet-B4 models trained on OpenEarthMap, LoveDA, DeepGlobe, and DynamicEarthNet with the same implementation details. The results obtained by the OpenEarthMap model demonstrate fine spatial details and semantically consistent mapping with the Chesapeake Bay land cover map. Note that the maps of the DynamicEarthNet model were obtained with the original 1m GSD because they yielded better results than those processed with 3m GSD. See Figure~\ref{fig:chesapeake2} for a comparison of mapping results with the DynamicEarthNet model using images at different GSDs (i.e., 0.5cm, 1m, and 3m) for inference.
	
\begin{figure*}[!t]
		\centering
		\begin{minipage}[b]{1\textwidth}
			\includegraphics[width=\linewidth]{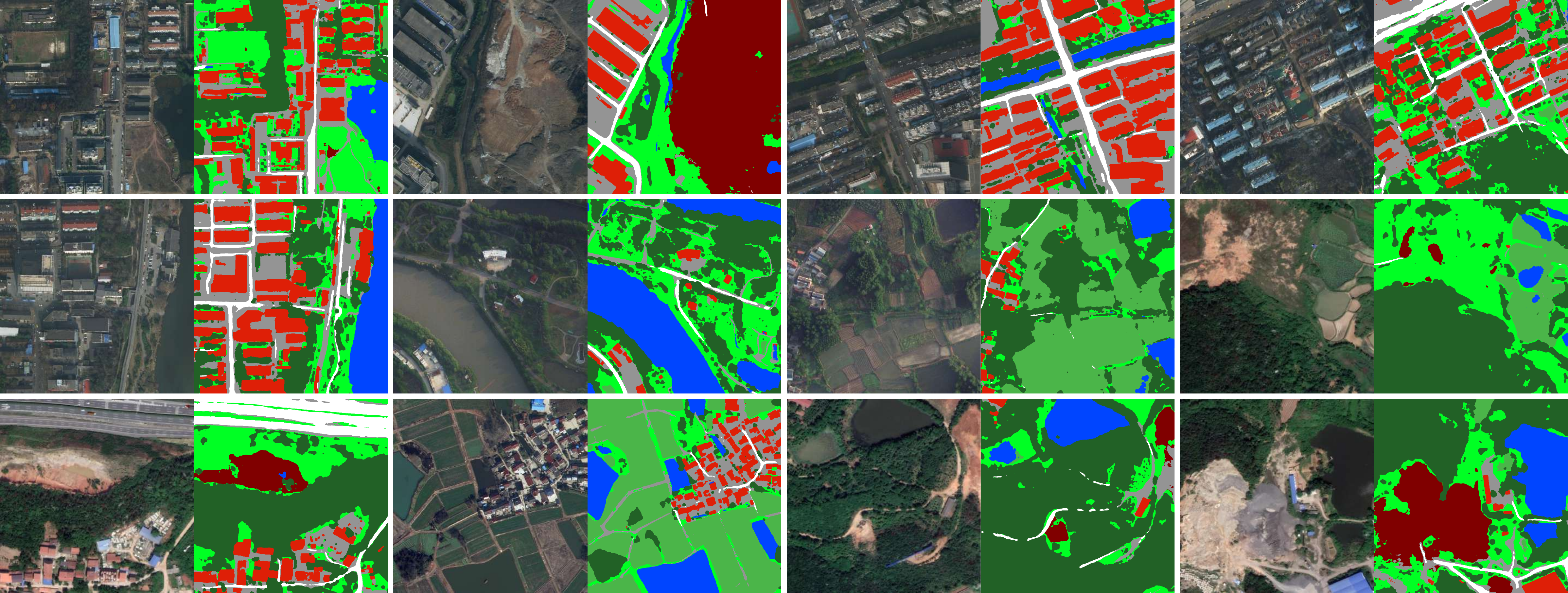}
			\vspace{-3mm}
			\caption{Out-of-sample mapping examples of LoveDA from China.}
			\label{fig:loveda}
			\vspace{10mm}
		\end{minipage}\hfil
		\begin{minipage}[b]{1\textwidth}
			\includegraphics[width=\linewidth]{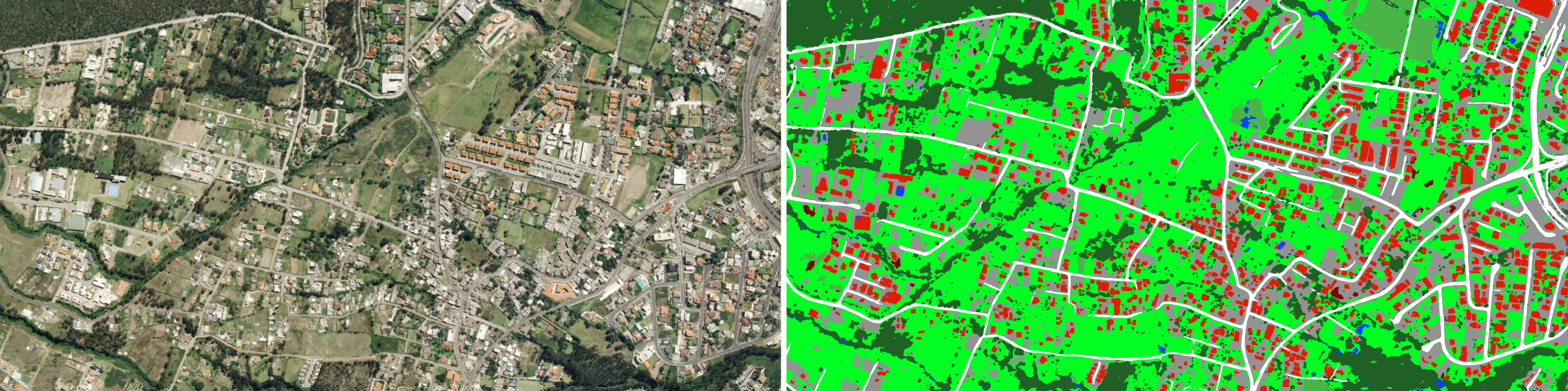}
			\vspace{-3mm}
			\caption{Out-of-sample mapping example of SIGTIERRAS from Ecuador.}
			\label{fig:sigtierras}
			\vspace{10mm}
		\end{minipage}\hfil
		\begin{minipage}[b]{1\textwidth}
			\includegraphics[width=\linewidth]{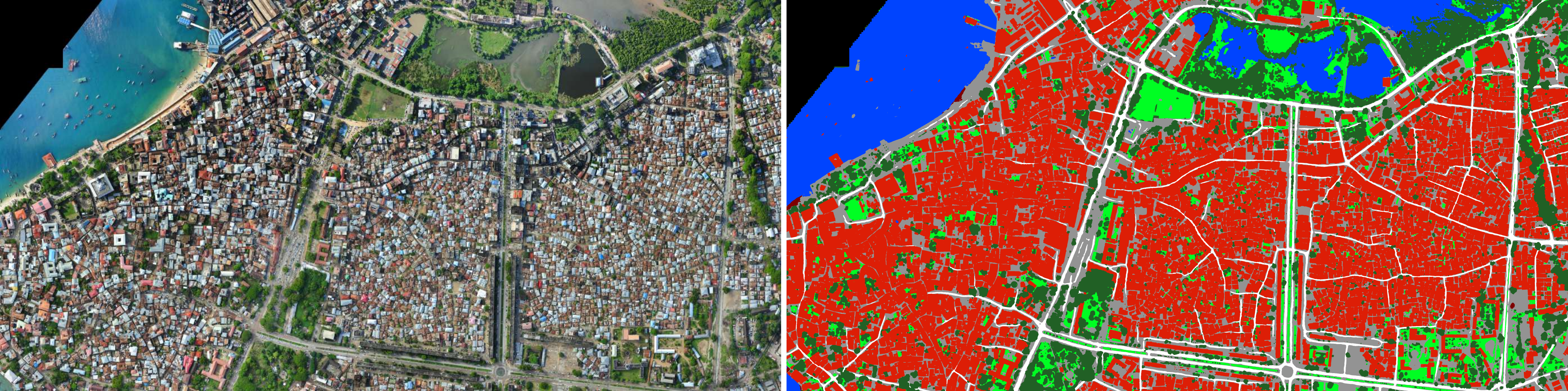}
			\vspace{-3mm}
			\caption{Out-of-sample mapping example of ZMI from Tanzania.}
			\label{fig:zmi}
		\end{minipage}
	\end{figure*}

	Furthermore, we demonstrate visual results of land cover mapping for out-of-sample images from France (MiniFrance~\cite{castillo2021semi}), China (LoveDA~\cite{wang2021loveda}), Ecuador (SIGTIERRAS\footnote{http://www.sigtierras.gob.ec/}), and Tanzania (Zanzibar Mapping Initiative (ZMI)\footnote{http://www.zanzibarmapping.org/}) in Figures~\ref{fig:minifrance}, \ref{fig:loveda}, \ref{fig:sigtierras}, and \ref{fig:zmi}, respectively. These land cover maps were obtained by a U-Net-EfficientNet-B4 trained on the OpenEarthMap dataset. One can observe that the results are at a reasonably high resolution. Considering the fact that these images are not included in the OpenEarthMap dataset, the results further support the generalization capability of the OpenEarthMap model.

	\section{Attribution of Source Data}
	Table~\ref{tab:attribution} summarizes attribution of source data for 97 regions in OpenEarthMap. Our label data are provided under the same license as the original RGB images, which varies with each source dataset. Label data for regions where the original RGB images are in the public domain or where the license is not explicitly stated are licensed under a \href{https://creativecommons.org/licenses/by-nc-sa/4.0/}{Creative Commons Attribution-NonCommercial-ShareAlike 4.0 International License}.

	\begin{table*}[t!]
		\caption{Attribution of source data of the 97 regions in OpenEarthMap.}
		\label{tab:attribution}
		\vspace{-5mm}
		\begin{center}
			\scalebox{0.85}{
				\tiny
				\begin{tabular}{p{0.1\textwidth}p{0.1\textwidth}p{0.07\textwidth}p{0.35\textwidth}p{0.3\textwidth}p{0.08\textwidth}}
					\hline\hline
					Region & Country & Source data & URL & Provider & License \\
					\hline
					Christchurch & New Zealand & AIRS & https://www.airs-dataset.com/ & Land Information of New Zealand & \href{https://creativecommons.org/licenses/by/4.0/}{CC BY 4.0} \\ 
					Aachen & Germany & GeoNRW & https://www.opengeodata.nrw.de/produkte/ & German state North Rhine-Westphalia & \href{https://www.govdata.de/dl-de/by-2-0}{DL-DE-BY-2.0} \\ 
					Bielefeld & Germany & GeoNRW & https://www.opengeodata.nrw.de/produkte/ & German state North Rhine-Westphalia & \href{https://www.govdata.de/dl-de/by-2-0}{DL-DE-BY-2.0} \\ 
					Dortmunt & Germany & GeoNRW & https://www.opengeodata.nrw.de/produkte/ & German state North Rhine-Westphalia & \href{https://www.govdata.de/dl-de/by-2-0}{DL-DE-BY-2.0} \\ 
					Dusseldorf & Germany & GeoNRW & https://www.opengeodata.nrw.de/produkte/ & German state North Rhine-Westphalia & \href{https://www.govdata.de/dl-de/by-2-0}{DL-DE-BY-2.0} \\ 
					Koeln & Germany & GeoNRW & https://www.opengeodata.nrw.de/produkte/ & German state North Rhine-Westphalia & \href{https://www.govdata.de/dl-de/by-2-0}{DL-DE-BY-2.0} \\ 
					Muenster & Germany & GeoNRW & https://www.opengeodata.nrw.de/produkte/ & German state North Rhine-Westphalia & \href{https://www.govdata.de/dl-de/by-2-0}{DL-DE-BY-2.0} \\ 
					Chisinau & Moldova & HTCD & https://map.openaerialmap.org & Lightcyphers & \href{https://creativecommons.org/licenses/by/4.0/}{CC BY 4.0} \\ 
					Tyrol & Austria & Inria & https://project.inria.fr/aerialimagelabeling/ & Tyrol & Public domain \\ 
					Vienna & Austria & Inria & https://project.inria.fr/aerialimagelabeling/ & Vienna & Public domain \\ 
					Austin & USA & Inria & https://project.inria.fr/aerialimagelabeling/ & USGS & Public domain \\ 
					Chicago & USA & Inria & https://project.inria.fr/aerialimagelabeling/ & USGS & Public domain \\ 
					Kitsap & USA & Inria & https://project.inria.fr/aerialimagelabeling/ & USGS & Public domain \\ 
					Dolnoslaskie & Poland & Landcover.ai & https://landcover.ai.linuxpolska.com/ & Head Office of Geodesy and Cartography & \href{https://creativecommons.org/licenses/by-nc-sa/4.0/}{CC BY-NC-SA 4.0} \\ 
					Kujawsko-pomorskie & Poland & Landcover.ai & https://landcover.ai.linuxpolska.com/ & Head Office of Geodesy and Cartography & \href{https://creativecommons.org/licenses/by-nc-sa/4.0/}{CC BY-NC-SA 4.0} \\ 
					Lodzkie & Poland & Landcover.ai & https://landcover.ai.linuxpolska.com/ & Head Office of Geodesy and Cartography & \href{https://creativecommons.org/licenses/by-nc-sa/4.0/}{CC BY-NC-SA 4.0} \\ 
					Lubuskie & Poland & Landcover.ai & https://landcover.ai.linuxpolska.com/ & Head Office of Geodesy and Cartography & \href{https://creativecommons.org/licenses/by-nc-sa/4.0/}{CC BY-NC-SA 4.0} \\ 
					Malopolskie & Poland & Landcover.ai & https://landcover.ai.linuxpolska.com/ & Head Office of Geodesy and Cartography & \href{https://creativecommons.org/licenses/by-nc-sa/4.0/}{CC BY-NC-SA 4.0} \\ 
					Mazowieckie & Poland & Landcover.ai & https://landcover.ai.linuxpolska.com/ & Head Office of Geodesy and Cartography & \href{https://creativecommons.org/licenses/by-nc-sa/4.0/}{CC BY-NC-SA 4.0} \\ 
					Podkarpackie & Poland & Landcover.ai & https://landcover.ai.linuxpolska.com/ & Head Office of Geodesy and Cartography & \href{https://creativecommons.org/licenses/by-nc-sa/4.0/}{CC BY-NC-SA 4.0} \\ 
					Podlaskie & Poland & Landcover.ai & https://landcover.ai.linuxpolska.com/ & Head Office of Geodesy and Cartography & \href{https://creativecommons.org/licenses/by-nc-sa/4.0/}{CC BY-NC-SA 4.0} \\ 
					Pomorskie & Poland & Landcover.ai & https://landcover.ai.linuxpolska.com/ & Head Office of Geodesy and Cartography & \href{https://creativecommons.org/licenses/by-nc-sa/4.0/}{CC BY-NC-SA 4.0} \\ 
					Slaskie & Poland & Landcover.ai & https://landcover.ai.linuxpolska.com/ & Head Office of Geodesy and Cartography & \href{https://creativecommons.org/licenses/by-nc-sa/4.0/}{CC BY-NC-SA 4.0} \\ 
					Swietokrzyskie & Poland & Landcover.ai & https://landcover.ai.linuxpolska.com/ & Head Office of Geodesy and Cartography & \href{https://creativecommons.org/licenses/by-nc-sa/4.0/}{CC BY-NC-SA 4.0} \\ 
					Warminsko-mazurskie & Poland & Landcover.ai & https://landcover.ai.linuxpolska.com/ & Head Office of Geodesy and Cartography & \href{https://creativecommons.org/licenses/by-nc-sa/4.0/}{CC BY-NC-SA 4.0} \\ 
					Wielkopolskie & Poland & Landcover.ai & https://landcover.ai.linuxpolska.com/ & Head Office of Geodesy and Cartography & \href{https://creativecommons.org/licenses/by-nc-sa/4.0/}{CC BY-NC-SA 4.0} \\ 
					Zachodniopomorskie & Poland & Landcover.ai & https://landcover.ai.linuxpolska.com/ & Head Office of Geodesy and Cartography & \href{https://creativecommons.org/licenses/by-nc-sa/4.0/}{CC BY-NC-SA 4.0} \\ 
					Ngaoundere & Cameroon & Open Cities AI & https://www.drivendata.org/competitions/60/building-segmentation-disaster-resilience/ & Global Facility for Disaster Reduction and Recovery & \href{https://creativecommons.org/licenses/by/4.0/}{CC BY 4.0} \\ 
					Kinshasa & Congo & Open Cities AI & https://www.drivendata.org/competitions/60/building-segmentation-disaster-resilience/ & Global Facility for Disaster Reduction and Recovery & \href{https://creativecommons.org/licenses/by/4.0/}{CC BY 4.0} \\ 
					Pointenoire & Congo & Open Cities AI & https://www.drivendata.org/competitions/60/building-segmentation-disaster-resilience/ & Global Facility for Disaster Reduction and Recovery & \href{https://creativecommons.org/licenses/by/4.0/}{CC BY 4.0} \\ 
					Accra & Ghana & Open Cities AI & https://www.drivendata.org/competitions/60/building-segmentation-disaster-resilience/ & Global Facility for Disaster Reduction and Recovery & \href{https://creativecommons.org/licenses/by/4.0/}{CC BY 4.0} \\ 
					Monrovia & Liberia & Open Cities AI & https://www.drivendata.org/competitions/60/building-segmentation-disaster-resilience/ & Global Facility for Disaster Reduction and Recovery & \href{https://creativecommons.org/licenses/by/4.0/}{CC BY 4.0} \\ 
					Niamey & Niger & Open Cities AI & https://www.drivendata.org/competitions/60/building-segmentation-disaster-resilience/ & Global Facility for Disaster Reduction and Recovery & \href{https://creativecommons.org/licenses/by/4.0/}{CC BY 4.0} \\ 
					Mahe & Seychelles & Open Cities AI & https://www.drivendata.org/competitions/60/building-segmentation-disaster-resilience/ & Global Facility for Disaster Reduction and Recovery & \href{https://creativecommons.org/licenses/by/4.0/}{CC BY 4.0} \\ 
					Dar es salaam & Tanzania & Open Cities AI & https://www.drivendata.org/competitions/60/building-segmentation-disaster-resilience/ & Global Facility for Disaster Reduction and Recovery & \href{https://creativecommons.org/licenses/by/4.0/}{CC BY 4.0} \\ 
					Dar es salaam & Tanzania & Open Cities AI & https://www.drivendata.org/competitions/60/building-segmentation-disaster-resilience/ & Global Facility for Disaster Reduction and Recovery & \href{https://creativecommons.org/licenses/by/4.0/}{CC BY 4.0} \\ 
					Zanzibar & Tanzania & Open Cities AI & https://www.drivendata.org/competitions/60/building-segmentation-disaster-resilience/ & Global Facility for Disaster Reduction and Recovery & \href{https://creativecommons.org/licenses/by/4.0/}{CC BY 4.0} \\ 
					Kampala & Uganda & Open Cities AI & https://www.drivendata.org/competitions/60/building-segmentation-disaster-resilience/ & Global Facility for Disaster Reduction and Recovery & \href{https://creativecommons.org/licenses/by/4.0/}{CC BY 4.0} \\ 
					Buenos aires & Argentina & Open data & https://map.openaerialmap.org & Municipalidad de Pergamino & \href{https://creativecommons.org/licenses/by/4.0/}{CC BY 4.0} \\ 
					Rosario & Argentina & Open data & https://map.openaerialmap.org & Julia Faraudello & \href{https://creativecommons.org/licenses/by/4.0/}{CC BY 4.0} \\ 
					Melbourne & Australia & Open data & https://map.openaerialmap.org & City of Melbourne & \href{https://creativecommons.org/licenses/by/4.0/}{CC BY 4.0} \\ 
					Cox's bazar & Bangladesh & Open data & https://map.openaerialmap.org & IOM Bangladesh - Needs and Population Monitoring (NPM) Drone & \href{https://creativecommons.org/licenses/by/4.0/}{CC BY 4.0} \\ 
					Cox's bazar & Bangladesh & Open data & https://map.openaerialmap.org & IOM Bangladesh - Needs and Population Monitoring (NPM) Drone & \href{https://creativecommons.org/licenses/by/4.0/}{CC BY 4.0} \\ 
					Dhaka & Bangladesh & Open data & https://map.openaerialmap.org & AIGEO Center & \href{https://creativecommons.org/licenses/by/4.0/}{CC BY 4.0} \\ 
					Santiago & Chile & Open data & https://map.openaerialmap.org & SECTRA & \href{https://creativecommons.org/licenses/by/4.0/}{CC BY 4.0} \\ 
					Bogota & Colombia & Open data & https://map.openaerialmap.org & Maptime Bogota & \href{https://creativecommons.org/licenses/by/4.0/}{CC BY 4.0} \\ 
					Svaneti & Georgia & Open data & https://map.openaerialmap.org & Transcaucasian Trail Association & \href{https://creativecommons.org/licenses/by/4.0/}{CC BY 4.0} \\ 
					Accra & Ghana & Open data & https://map.openaerialmap.org & Environmental Protection Agency Ghana & \href{https://creativecommons.org/licenses/by/4.0/}{CC BY 4.0} \\ 
					Western & Ghana & Open data & https://map.openaerialmap.org & UMaT YouthMappers & \href{https://creativecommons.org/licenses/by/4.0/}{CC BY 4.0} \\ 
					Al qurnah & Iraq & Open data & https://map.openaerialmap.org & Bilal Koç & \href{https://creativecommons.org/licenses/by/4.0/}{CC BY 4.0} \\ 
					Kyoto & Japan & Open data & https://maps.gsi.go.jp/development/ichiran.html & Geospatial Information Authority of Japan &  \\ 
					Tokyo & Japan & Open data & https://maps.gsi.go.jp/development/ichiran.html & Geospatial Information Authority of Japan &  \\ 
					Monrovia & Liberia & Open data & https://map.openaerialmap.org & Uhurulabs & \href{https://creativecommons.org/licenses/by/4.0/}{CC BY 4.0} \\ 
					Dowa & Malawi & Open data & https://map.openaerialmap.org & MapMalawi & \href{https://creativecommons.org/licenses/by/4.0/}{CC BY 4.0} \\ 
					Ulaanbaatar & Mongolia & Open data & https://map.openaerialmap.org & City of Ulaanbaatar and Asia Foundation & \href{https://creativecommons.org/licenses/by/4.0/}{CC BY 4.0} \\ 
					Maputo & Mozambique & Open data & https://map.openaerialmap.org & MapeandoMeuBairro & \href{https://creativecommons.org/licenses/by/4.0/}{CC BY 4.0} \\ 
					Abancay & Peru & Open data & https://sigrid.cenepred.gob.pe & National Center for Disaster Risk Estimation, Prevention and Reduction &  \\ 
					Chiclayo & Peru & Open data & https://sigrid.cenepred.gob.pe & National Center for Disaster Risk Estimation, Prevention and Reduction &  \\ 
					Chincha & Peru & Open data & https://sigrid.cenepred.gob.pe & National Center for Disaster Risk Estimation, Prevention and Reduction &  \\ 
					Ica & Peru & Open data & https://sigrid.cenepred.gob.pe & National Center for Disaster Risk Estimation, Prevention and Reduction &  \\ 
					Lambayeque & Peru & Open data & https://sigrid.cenepred.gob.pe & National Center for Disaster Risk Estimation, Prevention and Reduction &  \\ 
					Lima & Peru & Open data & https://sigrid.cenepred.gob.pe & National Center for Disaster Risk Estimation, Prevention and Reduction &  \\ 
					Pisco & Peru & Open data & https://sigrid.cenepred.gob.pe & National Center for Disaster Risk Estimation, Prevention and Reduction &  \\ 
					Piura & Peru & Open data & https://sigrid.cenepred.gob.pe & National Center for Disaster Risk Estimation, Prevention and Reduction &  \\ 
					Sechura & Peru & Open data & https://sigrid.cenepred.gob.pe & National Center for Disaster Risk Estimation, Prevention and Reduction &  \\ 
					Viru & Peru & Open data & https://sigrid.cenepred.gob.pe & National Center for Disaster Risk Estimation, Prevention and Reduction &  \\ 
					Baybay & Philippines & Open data & https://map.openaerialmap.org & SkyEye & \href{https://creativecommons.org/licenses/by/4.0/}{CC BY 4.0} \\ 
					San tome & Sao Tome and Principe & Open data & https://map.openaerialmap.org & Drones Adventures & \href{https://creativecommons.org/licenses/by/4.0/}{CC BY 4.0} \\ 
					Chiangmai & Tailand & Open data & https://map.openaerialmap.org & UR Field Lab Chiang Mai & \href{https://creativecommons.org/licenses/by/4.0/}{CC BY 4.0} \\ 
					Lohur & Tajikistan & Open data & https://map.openaerialmap.org & FAZO Institute, Dushanbe, Tajikistan & \href{https://creativecommons.org/licenses/by/4.0/}{CC BY 4.0} \\ 
					Kagera & Tanzania & Open data & https://map.openaerialmap.org & WeRobotics & \href{https://creativecommons.org/licenses/by/4.0/}{CC BY 4.0} \\ 
					Zanzibar & Tanzania & Open data & https://map.openaerialmap.org & Commission for Lands and Revolutionary Government of Zanzibar & \href{https://creativecommons.org/licenses/by/4.0/}{CC BY 4.0} \\ 
					Tonga & Tonga & Open data & https://map.openaerialmap.org & World Bank, V-TOL Aerospace & \href{https://creativecommons.org/licenses/by/4.0/}{CC BY 4.0} \\ 
					Soriano & Uruguay & Open data & https://map.openaerialmap.org & IntelDrones SRL & \href{https://creativecommons.org/licenses/by/4.0/}{CC BY 4.0} \\ 
					Rio & Brazil & SpaceNet & https://spacenet.ai/spacenet-buildings-dataset-v1/ & Maxar & \href{https://creativecommons.org/licenses/by-sa/4.0/}{CC BY-SA 4.0} \\ 
					Shanghai & China & SpaceNet & https://spacenet.ai/spacenet-buildings-dataset-v2/ & Maxar & \href{https://creativecommons.org/licenses/by-sa/4.0/}{CC BY-SA 4.0} \\ 
					Paris & France & SpaceNet & https://spacenet.ai/spacenet-buildings-dataset-v2/ & Maxar & \href{https://creativecommons.org/licenses/by-sa/4.0/}{CC BY-SA 4.0} \\ 
					Rotterdam & Netherlands & SpaceNet & https://spacenet.ai/sn6-challenge/ & Maxar & \href{https://creativecommons.org/licenses/by-sa/4.0/}{CC BY-SA 4.0} \\ 
					Khartoum & Sudan & SpaceNet & https://spacenet.ai/spacenet-buildings-dataset-v2/ & Maxar & \href{https://creativecommons.org/licenses/by-sa/4.0/}{CC BY-SA 4.0} \\ 
					Vegas & USA & SpaceNet & https://spacenet.ai/spacenet-buildings-dataset-v2/ & Maxar & \href{https://creativecommons.org/licenses/by-sa/4.0/}{CC BY-SA 4.0} \\ 
					Adelaide & Australia & xBD & https://xview2.org/ & Maxar & \href{https://creativecommons.org/licenses/by-nc-sa/4.0/}{CC BY-NC-SA 4.0} \\ 
					El rodeo & Guatemala & xBD & https://xview2.org/ & Maxar & \href{https://creativecommons.org/licenses/by-nc-sa/4.0/}{CC BY-NC-SA 4.0} \\ 
					Jeremie & Haiti & xBD & https://xview2.org/ & Maxar & \href{https://creativecommons.org/licenses/by-nc-sa/4.0/}{CC BY-NC-SA 4.0} \\ 
					Les-cayes & Haiti & xBD & https://xview2.org/ & Maxar & \href{https://creativecommons.org/licenses/by-nc-sa/4.0/}{CC BY-NC-SA 4.0} \\ 
					Port-a-piment & Haiti & xBD & https://xview2.org/ & Maxar & \href{https://creativecommons.org/licenses/by-nc-sa/4.0/}{CC BY-NC-SA 4.0} \\ 
					Saint-louis-du-sud & Haiti & xBD & https://xview2.org/ & Maxar & \href{https://creativecommons.org/licenses/by-nc-sa/4.0/}{CC BY-NC-SA 4.0} \\ 
					Palu & Indonesia & xBD & https://xview2.org/ & Maxar & \href{https://creativecommons.org/licenses/by-nc-sa/4.0/}{CC BY-NC-SA 4.0} \\ 
					Labuhan & Malaysia & xBD & https://xview2.org/ & Maxar & \href{https://creativecommons.org/licenses/by-nc-sa/4.0/}{CC BY-NC-SA 4.0} \\ 
					Mexico city & Mexico & xBD & https://xview2.org/ & Maxar & \href{https://creativecommons.org/licenses/by-nc-sa/4.0/}{CC BY-NC-SA 4.0} \\ 
					Gorakhpur & Nepal & xBD & https://xview2.org/ & Maxar & \href{https://creativecommons.org/licenses/by-nc-sa/4.0/}{CC BY-NC-SA 4.0} \\ 
					Pedrogao grande & Portugal & xBD & https://xview2.org/ & Maxar & \href{https://creativecommons.org/licenses/by-nc-sa/4.0/}{CC BY-NC-SA 4.0} \\ 
					Houston & USA & xBD & https://xview2.org/ & Maxar & \href{https://creativecommons.org/licenses/by-nc-sa/4.0/}{CC BY-NC-SA 4.0} \\ 
					Joplin & USA & xBD & https://xview2.org/ & Maxar & \href{https://creativecommons.org/licenses/by-nc-sa/4.0/}{CC BY-NC-SA 4.0} \\ 
					Leilane estates & USA & xBD & https://xview2.org/ & Maxar & \href{https://creativecommons.org/licenses/by-nc-sa/4.0/}{CC BY-NC-SA 4.0} \\ 
					Little rock & USA & xBD & https://xview2.org/ & Maxar & \href{https://creativecommons.org/licenses/by-nc-sa/4.0/}{CC BY-NC-SA 4.0} \\ 
					Oklahoma & USA & xBD & https://xview2.org/ & Maxar & \href{https://creativecommons.org/licenses/by-nc-sa/4.0/}{CC BY-NC-SA 4.0} \\ 
					Panama city & USA & xBD & https://xview2.org/ & Maxar & \href{https://creativecommons.org/licenses/by-nc-sa/4.0/}{CC BY-NC-SA 4.0} \\ 
					Santa rosa & USA & xBD & https://xview2.org/ & Maxar & \href{https://creativecommons.org/licenses/by-nc-sa/4.0/}{CC BY-NC-SA 4.0} \\ 
					Thousand oaks & USA & xBD & https://xview2.org/ & Maxar & \href{https://creativecommons.org/licenses/by-nc-sa/4.0/}{CC BY-NC-SA 4.0} \\ 
					Tulsa & USA & xBD & https://xview2.org/ & Maxar & \href{https://creativecommons.org/licenses/by-nc-sa/4.0/}{CC BY-NC-SA 4.0} \\ 
					Tuscaloosa & USA & xBD & https://xview2.org/ & Maxar & \href{https://creativecommons.org/licenses/by-nc-sa/4.0/}{CC BY-NC-SA 4.0} \\ 
					Wallace & USA & xBD & https://xview2.org/ & Maxar & \href{https://creativecommons.org/licenses/by-nc-sa/4.0/}{CC BY-NC-SA 4.0} \\ 
					\hline\hline
			\end{tabular}}
			\vspace{-5mm}
		\end{center}
	\end{table*}


%

\end{document}